\PassOptionsToPackage{dvipsnames,table}{xcolor}
\documentclass[10pt,twocolumn,letterpaper]{article}
\usepackage{float}
\usepackage[pagenumbers]{cvpr} 
\usepackage{bbm}

\definecolor{cvprblue}{rgb}{0.21,0.49,0.74}
\usepackage[pagebackref,breaklinks,colorlinks,allcolors=cvprblue]{hyperref}
 \usepackage{multirow}
\def\paperID{8252} 
\def\confName{CVPR}
\def\confYear{2025}

\title{Learning Physics-Based Full-Body Human Reaching and Grasping \\
from Brief Walking References}

\author{
Yitang Li\footnotemark[1]~~\textsuperscript{1,2,5}~~
Mingxian Lin\footnotemark[1]~~\textsuperscript{2,4}~~\\
Zhuo Lin\textsuperscript{1}~~
Yipeng Deng\textsuperscript{1,2 }~~
Yue Cao\textsuperscript{1,2 }~~
Li Yi\footnotemark[2]~~\textsuperscript{1,3,2}
\smallskip\\
\textsuperscript{1}Tsinghua University~~
\textsuperscript{2}Shanghai Qi Zhi Institute~~\\
\textsuperscript{3}Shanghai Artificial Intelligence Laboratory~~
\textsuperscript{4}The University of Hong Kong~~
\textsuperscript{5}Galbot~
\\
\href{https://liyitang22.github.io/phys-reach-grasp/}{\textcolor{magenta}{https://liyitang22.github.io/phys-reach-grasp/}}
}
\usepackage{adjustbox}
\definecolor{best}{rgb}{0.96, 0.57, 0.58}
\definecolor{second}{rgb}{0.98, 0.78, 0.57}
\definecolor{third}{rgb}{1.0, 1.0, 0.56}

\begin{document}
\maketitle
\footnotetext[1]{Equal Contribution.}
\footnotetext[2]{Corresponding Author.}
\begin{abstract}
Existing motion generation methods based on MoCap data are often limited by data quality and coverage. In this work, we propose a framework that generates diverse, physically feasible full-body human reaching and grasping motions using only brief walking MoCap data. 
Based on the observation that walking data captures valuable movement patterns transferable across tasks and, on the other hand, the advanced kinematic methods can generate diverse grasping poses, which can then be interpolated into motions to serve as task-specific guidance. 
Our approach incorporates an active data generation strategy to maximize the utility of the generated motions, along with a local feature alignment mechanism that transfers natural movement patterns from walking data to enhance both the success rate and naturalness of the synthesized motions. By combining the fidelity and stability of natural walking with the flexibility and generalizability of task-specific generated data, our method demonstrates strong performance and robust adaptability in diverse scenes and with unseen objects.

\end{abstract}

\section{Introduction}

\label{sec:intro}
Human-object interactions are central to our engagement with the physical world. 
By naturally reaching for and grasping diverse objects in complex scenes, we enable rich, functional interactions for various tasks. 
Consequently, generating realistic, physics-based full-body motions for human reaching and grasping diverse objects in different scenes has broad applications in animation and AR/VR, and holds significant potential for advancing humanoid robotics.

Previous data-driven works\cite{ASE,AMP,CALM,controlvae,NPMP,physicsvae,PULSE,phc,ncp,phy} have shown promising results in interactive human motion generation, utilizing adversarial learning to replicate motions from a given dataset\cite{ASE,AMP,CALM} or employing explicit motion tracking to imitate real-world data\cite{controlvae,NPMP,physicsvae,PULSE,phc,ncp}. 
Although these methods produce high-quality motions, they tend to generate motions that closely follow the reference states~\cite{pmp}, remaining restricted by the limitations of the MoCap dataset.
However, human interactive skills are highly diverse, especially in complex reaching and grasping involving intricate object geometry and scene variations. 
Collecting motion capture data for such tasks is significantly time- and labor-intensive, often leading to limited motion samples and biased distributions. 
Therefore, instead of relying on extensive motion data collections, we take a bold step by exploring the generation of diverse full-body reaching and grasping motions using only brief, easily accessible walking references.

\begin{figure}[t]
    \centering
    \includegraphics[width=0.85\linewidth]{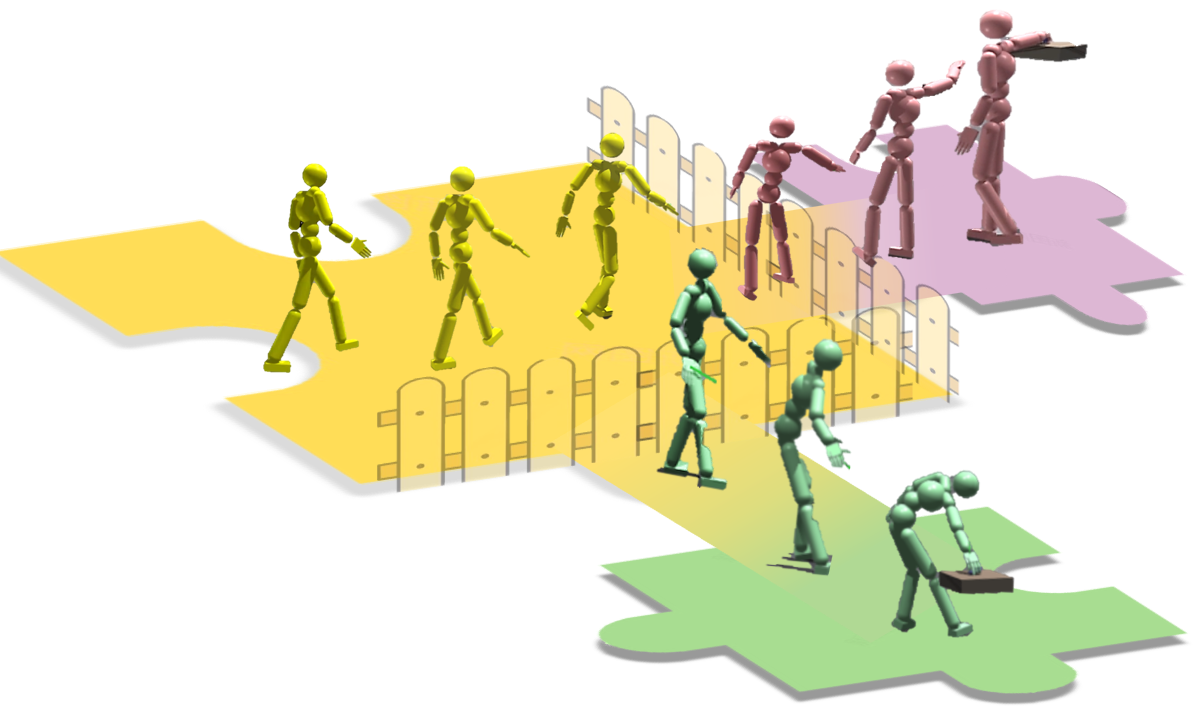}
    \caption{In this work, we design a framework that generates diverse, physically feasible full-body human reaching and grasping motions using only brief walking MoCap data.}
    \label{fig:enter-label}

\end{figure}

We observe that despite the dramatically different semantics and global movements between walking and full-body reaching and grasping, walking motions encompass diverse local motor patterns and balancing abilities (such as lifting the right hand while extending the left foot) that are universally applicable across different actions, providing a valuable motion pattern reference. 
On the other hand, recent advanced kinematic methods can utilize pose priors to generate diverse high-quality grasping poses conditioned on specific scenes and objects. 
These poses can be interpolated to form continuous motions. 
Though lacking physical guarantees and the natural patterns of human movement, they can also serve as useful task-specific guidance.

Based on the observations above, we propose to tackle the challenging motion synthesis problem by combining two data sources: limited, real MoCap walking data and flexible, task-adaptive, but unverified generated data. 
This integration faces two challenges: determining which tasks the generated data should target, and effectively combining real and generated data while avoiding artifacts in the generated data. 
For the first, we employ an active strategy to generate more data for tasks with worse performance to precisely address difficult scenes, maximizing the utility of the generated motions. 
For the second, we aim to learn natural patterns of human movement from real walking data and transfer them to new motion generation. 
A pilot study revealed that when passing a motion into a network, the real MoCap data(even with different semantics, like walking and reaching) tend to cluster in shallow network layers, while deeper layers tend to capture broader motion semantics. 
Based on this, we developed a feature alignment mechanism at shallow layers to regularize the motion space during the training on the augmented dataset, improving both the success rate and naturalness of the synthesized motions.

We conducted a series of experiments using our methods, and the results demonstrate that we can synthesize human-like reaching and grasping motions with a high success rate across diverse scenes and objects, all based solely on walking reference data. 
Further ablation studies reveal that both the active strategy and the feature alignment mechanism are crucial to our performance. 
The transferable local motion patterns play a significant role, not only in reducing artifacts but also in ensuring high stability when completing challenging grasping tasks.

In conclusion, our main contribution is the design of a framework that generates diverse, physically feasible full-body interactive human motions using only a limited amount of real MoCap walking data. 
We employ an active data generation strategy and a local feature alignment mechanism to leverage both the high fidelity and stability of natural walking motion data and the flexibility and generalizability of task-oriented generated data. 
Our approach demonstrates good performance and robust adaptability across diverse scenes and unseen objects.

\section{Related Works}
\label{sec:related}

\subsection{Human-Object Interactive Motion Synthesis}

Many existing works focus on full-body human-object interactive motion synthesis. Kinematic methods\cite{46_taheri2022goal,49_wu2022saga,23_li2024task} that use generative models for full-body grasps achieve high fidelity, but lack physical plausibility and depend heavily on the dataset. 
Recent approaches use physics simulations for human-object interaction \cite{5_chao2021learning,6_christen2022d,14_hassan2023synthesizing,25_luo2022embodied} but focus less on the modeling of the whole procedure including approaching, reaching and dexterous grasping. 
Similarly to us, Braun et al.\cite{phy} generate the entire reach and grasp motion but are limited to the human-object interaction dataset GRAB\cite{GRAB}. 
Omnigrasp\cite{omnigrasp} exhibits enhanced flexibility and generalizability, yet depends heavily on extensive training datasets like AMASS\cite{AMASS}. 
Unlike these methods, our work uses very limited motion data (walking) to synthesize high-fidelity motion sequences with a high success rate.

\subsection{Physics-based Motion Space Construction}
Recent advances leverage adversarial learning or motion-tracking objectives to create reusable representations of motion. 
ASE\cite{ASE} and CALM\cite{CALM} use discriminator and encoder rewards to map random noise to realistic behavior, which works well with small, specialized datasets. 
On the other hand, many works\cite{controlvae, NPMP, physicsvae, PULSE,ncp} employ motion tracking to potentially form a universal latent space. 
NCP\cite{ncp} learns this space by imitation with RL, whereas other model-based methods\cite{controlvae, physicsvae} incorporate a learned world model. 
PULSE\cite{PULSE} distills a motion representation from a pre-trained imitator\cite{phc} and a jointly learned prior. 
Although these methods show impressive results, they struggle to generate motions that lie far outside the distribution of their training data.

\section{Pilot Study}


In this section, we investigate whether real walking motions contain transferable features that can enhance full-body reaching and grasping movements. 
We employ a straightforward approach to examine this. 
Initially, we train a motion critic using brief walking MoCap data using data-driven motion synthesis techniques such as ASE\cite{ASE}. 
This critic is subsequently employed to assess two reaching motion datasets: one featuring authentic reaching movements and the other derived from interpolations between standing and grasping postures. 
This methodology allows us to identify any transferable features from walking to reaching and grasping, while also emphasizing the differences between real and interpolated motions.

\begin{figure}
    \centering
    \includegraphics[width=\linewidth]{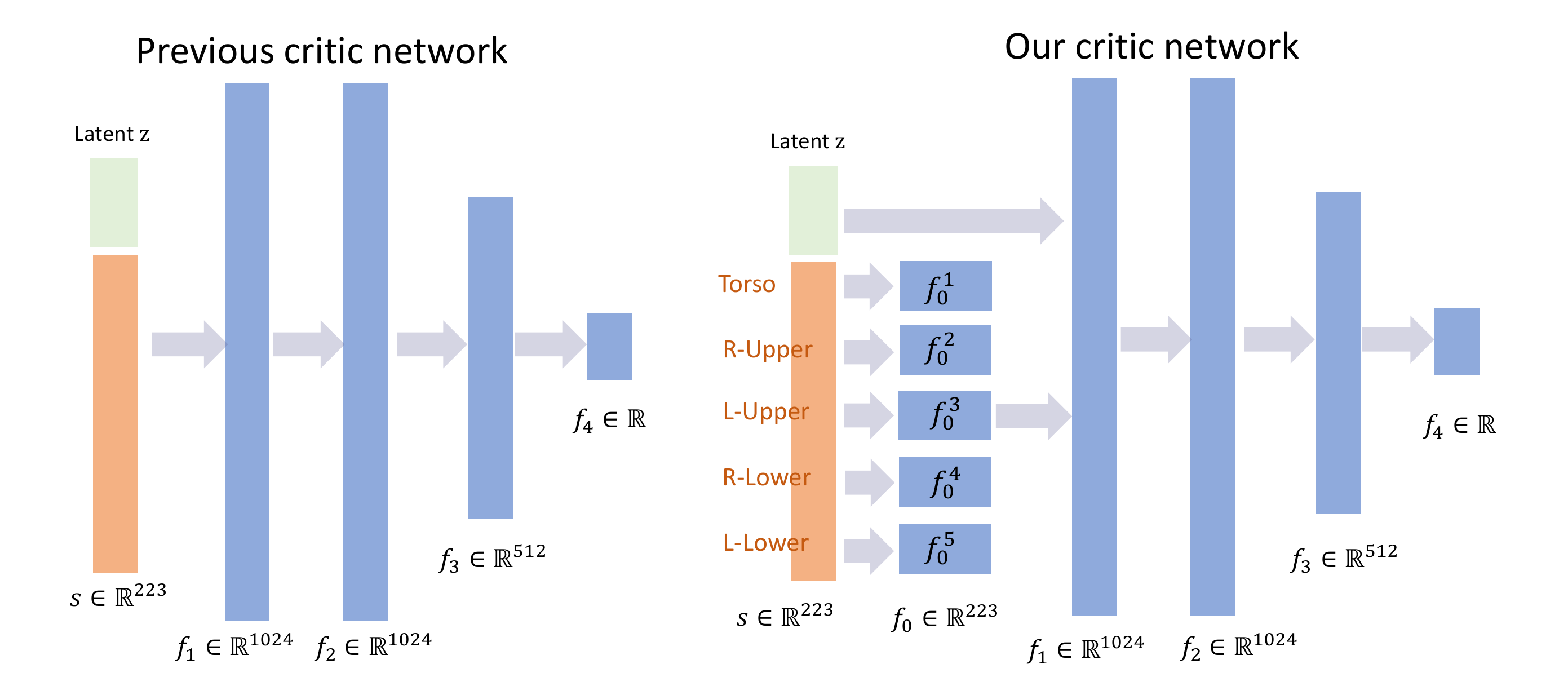}
    \caption{Comparison of our modified critic architecture.}
    \label{fig:Critic-architecture}
 
\end{figure}
\begin{table*}[h!t]
\small
\centering
\begin{tabular}{|c|ccccc|c|c|c|}
\hline
\multirow{2}{*}{FID Value}     & \multicolumn{5}{c|}{$f_0$}                                                                                              & \multirow{2}{*}{$f_1$} & \multirow{2}{*}{$f_2$} & \multirow{2}{*}{$f_3$} \\ \cline{2-6}
                               & \multicolumn{1}{c|}{$f_0^1$}  & \multicolumn{1}{c|}{$f_0^2$} & \multicolumn{1}{c|}{$f_0^3$} & \multicolumn{1}{c|}{$f_0^4$} & $f_0^5$ &                           &                           &                           \\ \hline
MoCap-Reach(train \& test)     & \multicolumn{1}{c|}{0.1750} & \multicolumn{1}{c|}{0.1960} & \multicolumn{1}{c|}{0.2092} & \multicolumn{1}{c|}{0.2452} & 0.0687 & 1.2870                    & 0.2414                     & 0.0344                   \\ \hline
MoCap-Reach \& MoCap-Walk      & \multicolumn{1}{c|}{2.3531}  & \multicolumn{1}{c|}{2.9465} & \multicolumn{1}{c|}{3.5363} & \multicolumn{1}{c|}{1.4978} & 0.6770 & 10.588                     & 4.0218                     & 1.3064                    \\ \hline
MoCap-Reach \& Generated-Reach & \multicolumn{1}{c|}{2.4671}       & \multicolumn{1}{c|}{8.5466}      & \multicolumn{1}{c|}{9.3327}      & \multicolumn{1}{c|}{1.5217}      &   0.9544    &     21.798                      &     6.3010                      &                  2.2653         \\ \hline
MoCap-Walk \& Generated-Reach  & \multicolumn{1}{c|}{2.9492}       & \multicolumn{1}{c|}{7.0355}      & \multicolumn{1}{c|}{11.340}      & \multicolumn{1}{c|}{1.6044}      &   0.9823    &              25.650             &     8.8653                      &                   3.4491        \\ \hline
\end{tabular}
\caption{\textbf{Quantitative Results of Pilot Study:} In the shallower layers, the FID difference between the MoCap data is relatively small, compared to that between the MoCap and generated datasets. As feature extraction moves to deeper layers, the FID value increases significantly compared to the reference FID between training and test datasets.}

\label{tab:pilot_study}
\end{table*}
\begin{figure*}[!ht]
    \centering
    \begin{subfigure}{0.26\textwidth}
        \centering
        \includegraphics[width=\linewidth]{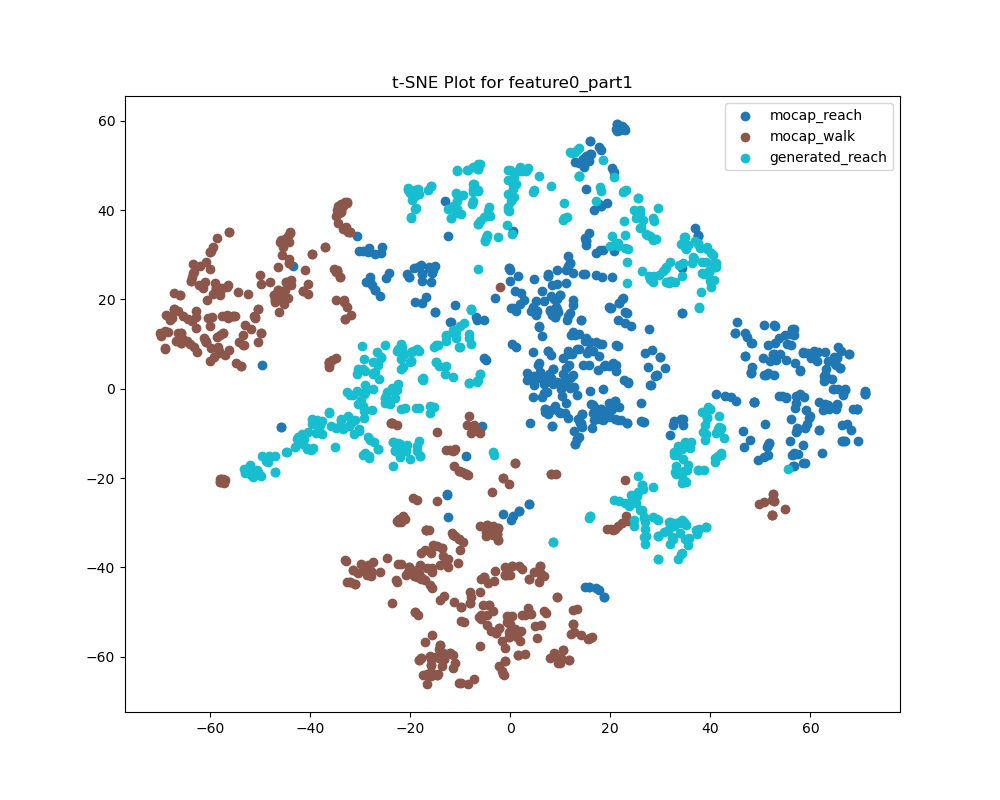}
    \end{subfigure}
    \hspace{-6mm}
    \begin{subfigure}{0.26\textwidth}
        \centering
\includegraphics[width=\linewidth]{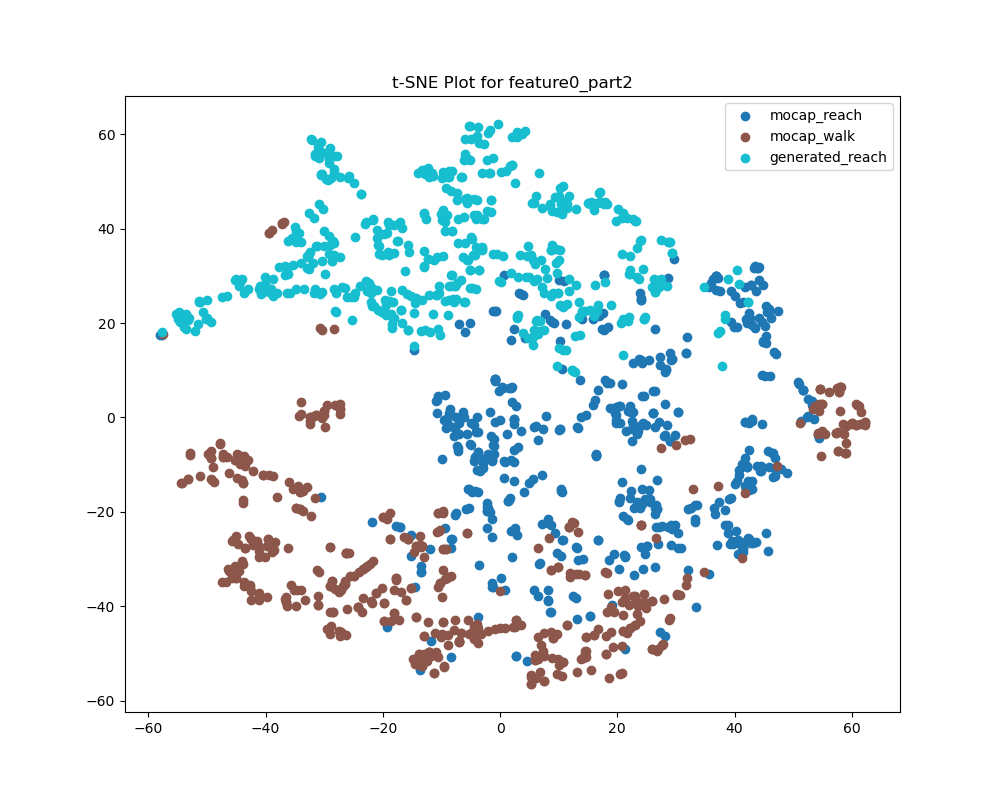}
    \end{subfigure}
        \hspace{-5.5mm}
    \begin{subfigure}{0.26\textwidth}
        \centering
        \includegraphics[width=\linewidth]{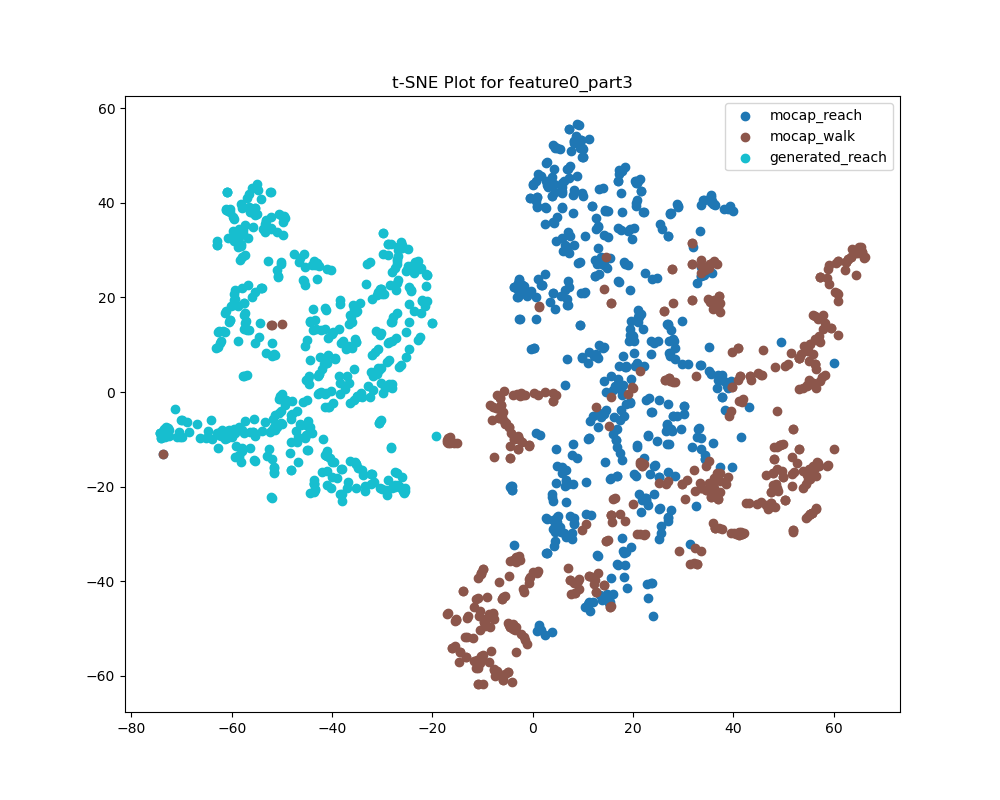}
    \end{subfigure}
        \hspace{-5.5mm}
    \begin{subfigure}{0.26\textwidth}
        \centering
        \includegraphics[width=\linewidth]{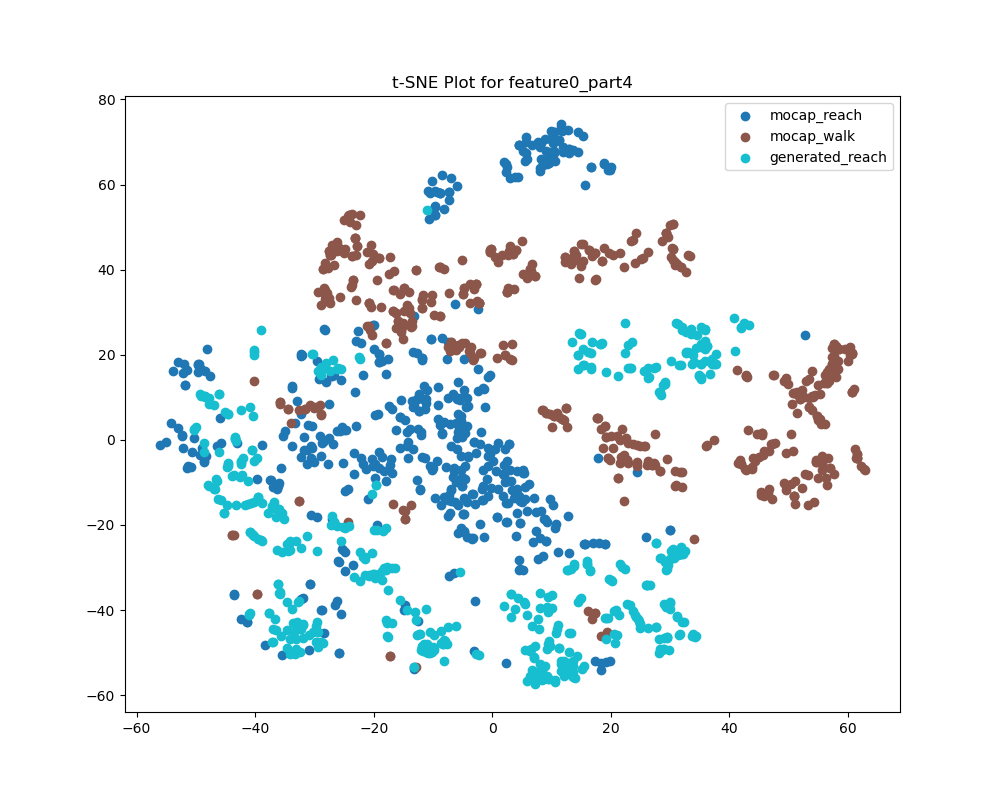}
    \end{subfigure}
    \begin{subfigure}{0.26\textwidth}
        \centering
        \includegraphics[width=\linewidth]{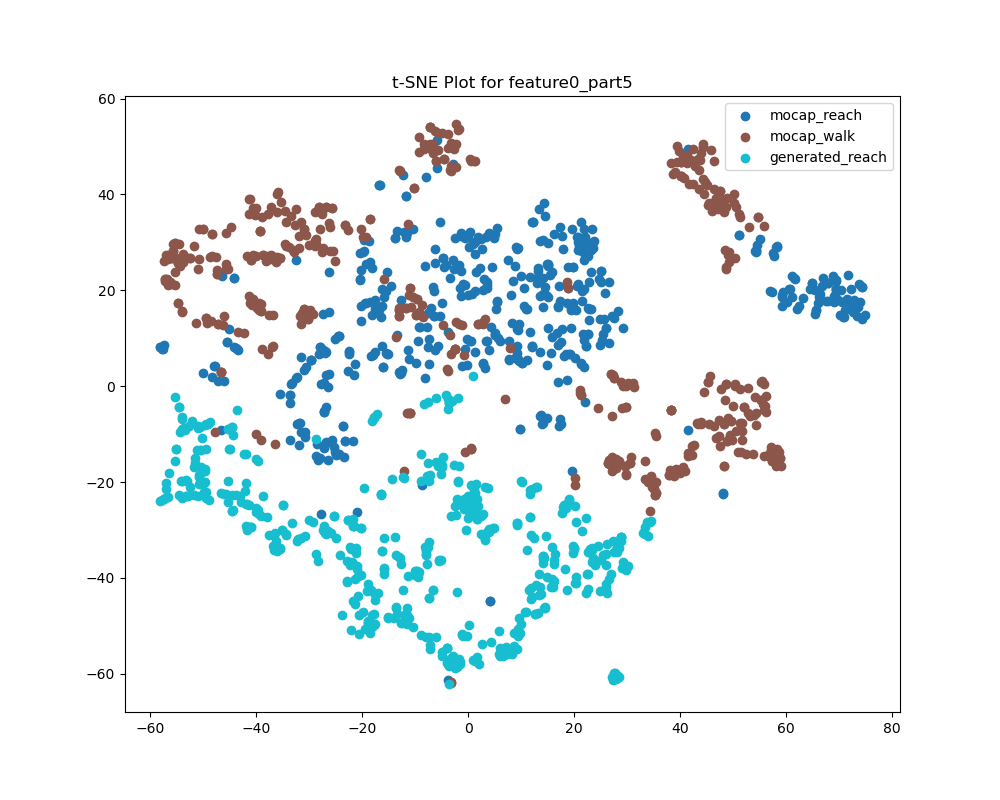}
    \end{subfigure}
          \hspace{-5.5mm}
    \begin{subfigure}{0.26\textwidth}
        \centering
        \includegraphics[width=\linewidth]{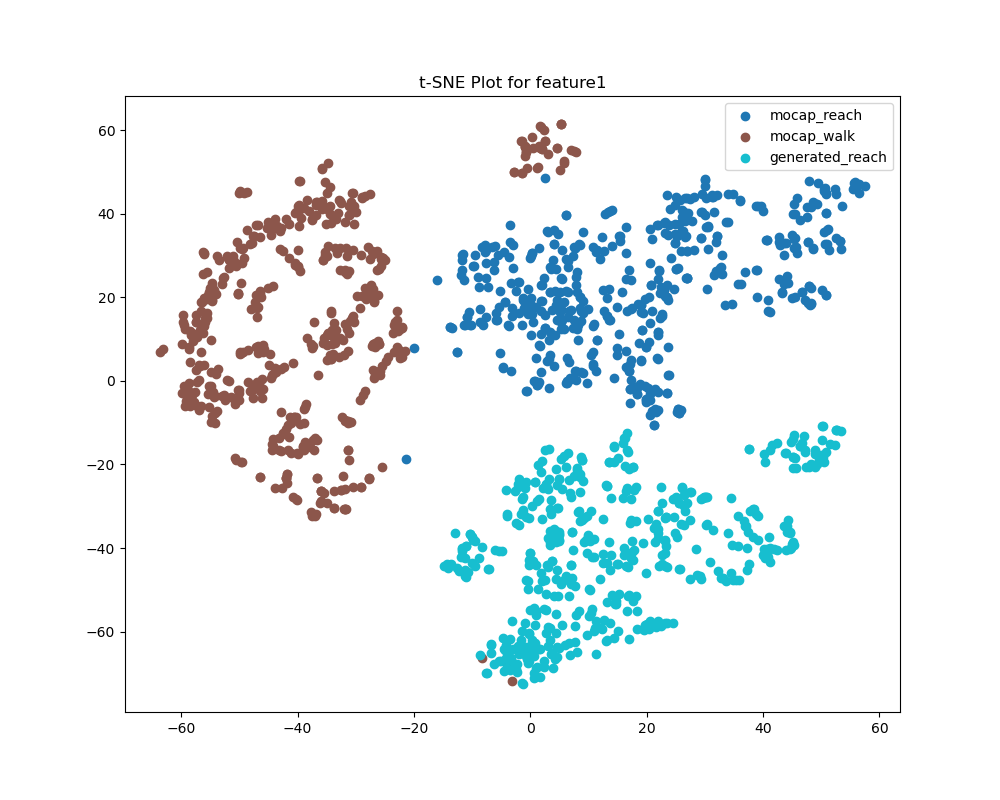}
    \end{subfigure}
         \hspace{-5.5mm}
    \begin{subfigure}{0.26\textwidth}
        \centering
        \includegraphics[width=\linewidth]{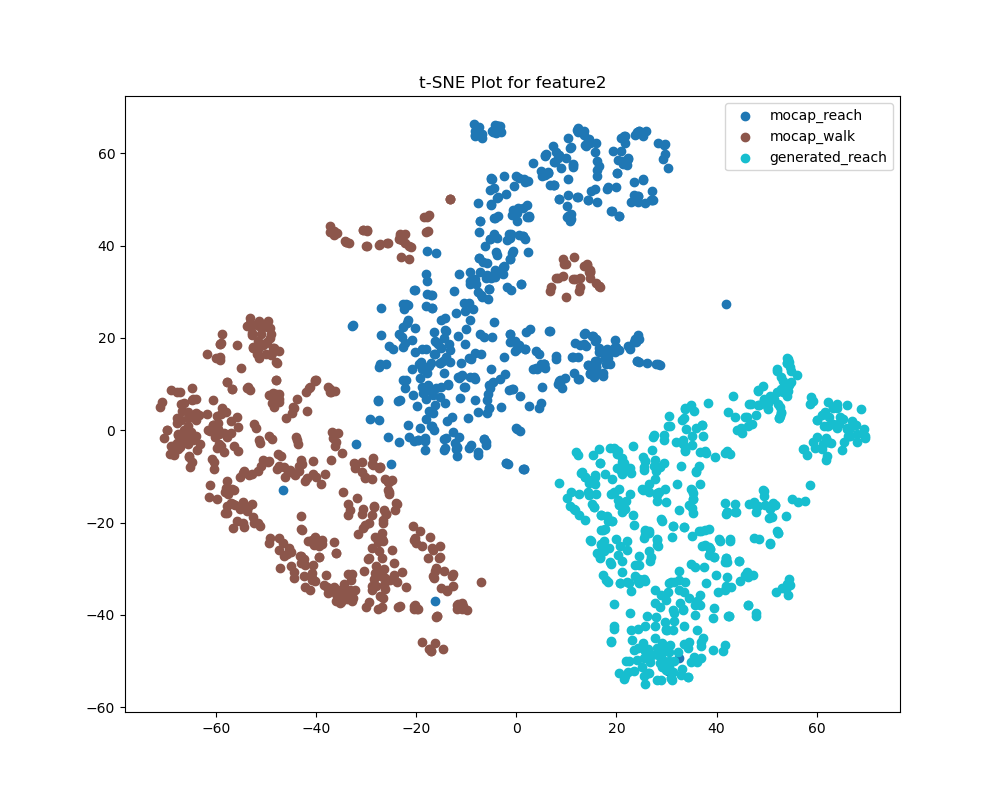}
    \end{subfigure}
           \hspace{-5.5mm}
    \begin{subfigure}{0.26\textwidth}
        \centering
        \includegraphics[width=\linewidth]{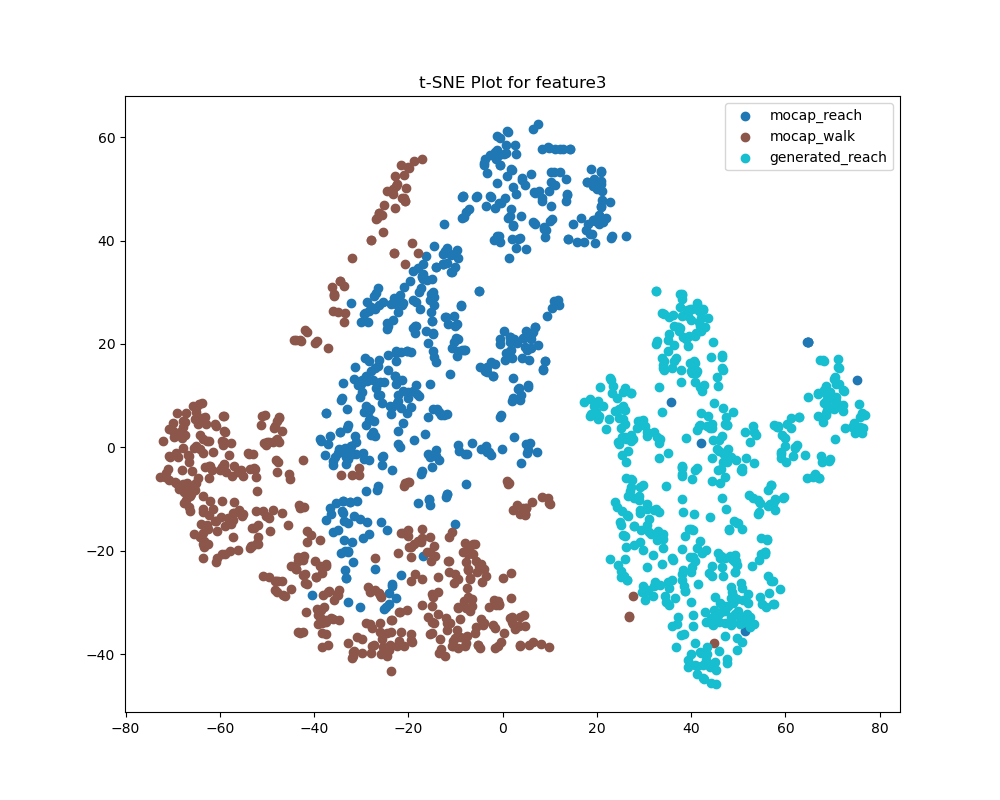}
    \end{subfigure}
    \caption{\textbf{t-SNE plots of features extracted at different levels of the critic network}: There is clear clustering within the MoCap data in shallow layers and this phenomenon is less evident in deeper layers.}

    \label{fig:pilot_study}
\end{figure*}

\subsection{Settings}

 Our training dataset for walking motions (MoCap-Walk) consisted of MoCap data sourced from AMASS~\cite{AMASS} and CMU~\cite{CMU}. 
Meanwhile, the dataset for reaching motions (MoCap-Reach) was derived from CIRCLE~\cite{circle}. 
Additionally, another dataset was produced via SLERP interpolation, transitioning from the standing pose to grasping poses as generated by FLEX~\cite{FLEX}.

In order to incorporate both global and local motion attributes into our analysis, we adjust the structure of the critic neural network, depicted in Figure~\ref{fig:Critic-architecture}. 
Initially, the network extracts segment-specific information, progressively blending these local features with the global ones. 
We categorize the input states into five distinct groups according to related body parts: (1) torso, (2) right upper limb, (3) left upper limb, (4) right lower limb, and (5) left lower limb. 
Each group is processed through a multi-layer perceptron (MLP), with the outputs concatenated before passing through further fully connected layers.

The critic trained on walking motion, is employed to assess the two additional datasets by extracting features from each layer. 
Following this, we measure the Fréchet Inception Distance (FID) between the resultant feature distributions and those of the training dataset. We also produce t-SNE plots to enhance the visualization of similarities.

\subsection{Results and Conclusions}

The results are presented in Table~\ref{tab:pilot_study} and Figure~\ref{fig:pilot_study}. We observe that, at features extracted from shallow layers, real data exhibits clear clustering (especially in $f_0$, except $f_0^1$) and lower FID values, despite differences in semantics(walking and grasping). In contrast, deeper features place greater emphasis on semantic information, such as whether two motions are similarly "walk-like", diminishing the clustering effect observed in shallow layers. Inspired by this key observation, we believe that the information extracted by shallow layers captures cross-task and cross-semantic cues about the essential characteristics of real MoCap data. The clustering motivates us to encourage new motions to align with the shallow-layer features of brief walking motion, serving as a form of regularization. We believe this alignment can help enhance motion realism and improve the stability in reaching and grasping.


\section{Methods}

\begin{figure*}[t]
    \centering
    \includegraphics[width=0.9\linewidth]{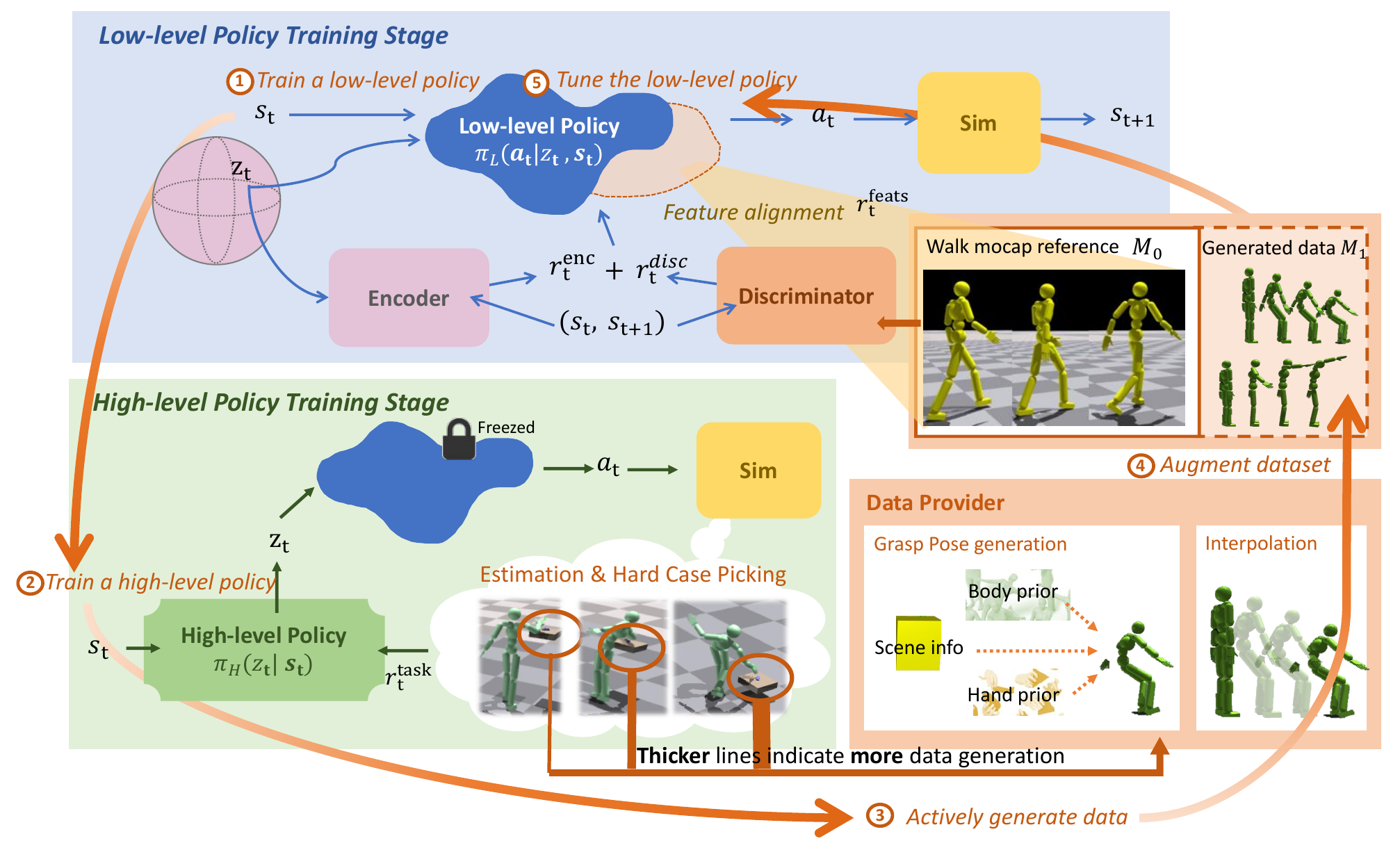}
    \caption{\textbf{Overview of our framework:} We propose a pipeline that generates diverse reaching and grasping motions using brief walk MoCap data through the multi-iteration training. In each iteration, with the imitation and discovery objective specified respectively by the discriminator and encoder, we first train a low-level policy $\pi_{L}(a|z,s)$ to map a latent variable $z$ to motions in the dataset. Next, using a task-specific reward, a high-level policy $\pi_H(z|s)$is trained to select $z$ to output actions for downstream tasks. After the first iteration, the motion space (represented by the low-level policy) contains only limited walking motions, restricting performance in challenging reaching and grasping tasks. To address this, we estimate the performance, identify hard cases, and actively generate interpolated data to augment the dataset. In subsequent iterations, we fine-tune the low-level policy on the augmented dataset to expand the motion space, using a feature alignment mechanism to regularize the output motions and provide an additional reward $r^{feats}$. Then, we can train the next-iteration high-level policy and this iterative process continues until we achieve satisfactory results.}

    \label{fig:pip}
\end{figure*}
To synthesize reaching and grasping motions from brief walking data, we first construct a latent motion skill space. We then train a policy to execute reaching and grasping tasks by leveraging skills from this latent space in a similar flavor to ASE~\cite{ASE}. To enhance the latent space with the necessary skills for reaching and grasping, we generate kinematic motions through interpolation between standing and grasping poses. This data is used to expand the latent motion space, while the shallow-layer skill features are still enforced to align with walking skills, ensuring the preservation of natural movement patterns.

As shown in Figure~\ref{fig:pip}, our pipeline can be viewed as a multi-iteration training process, where each iteration consists of two stages: a low-level policy training stage, in which we map latent variable $z$ to motions in the dataset, and a high-level policy training stage, in which the policy is trained to select $z$ to output motions for downstream tasks. More details will be presented in Section~\ref{low-level-space}~\ref{high-level-policy}. In the first iteration, the low-level policy is trained on a brief walking MoCap dataset $M_0$. The resulting motion space, limited to walking actions, fails to output actions to complete challenging reaching and grasping tasks. By assessing task performance, we identify challenging cases and actively generate additional data aligned with these cases to augment the dataset for the next iteration. The data generation pipeline and active strategy are discussed in Section~\ref{active}. In subsequent low-level training, we tune the low-level policy on the augmented dataset and incorporate a feature alignment mechanism by adding a reward $r^{feats}$ to encourage motions that retain similar shallow-layer features as walking. This mechanism is explained further in Section~\ref{feats}. This active training process is performed iteratively until satisfactory results are achieved.



\subsection{Low-level policy training}
~\label{low-level-space}
In this stage, we want to train a low-level policy $\pi(a|s,z)$ which maps a latent variable $z$ to a motion in the dataset $M$. The policy is trained with two objectives: an adversarial imitation objective specified by a discriminator $D(s_t,s_{t+1})$ and a skill discovery objective specified by an encoder $q(z|s_t,s_{t+1})$. More specifically, the reward can be expressed as:
 \begin{equation}
    r_t = -\log(1 - D(s_t,s_{t+1})) + \beta \log q(z|s_t,s_{t+1}) 
 \end{equation}
where $\beta$ is a constant to balance the two objectives. The policy is trained using the PPO algorithm that uses an actor to represent the policy and a critic conditioned on latent to estimate the value resembling all rewards. This critic also served as a motion feature extractor in our framework.


\subsection{High-level Policy Training}

\label{high-level-policy}

The high-level policy selects latent variable $z$ and passes it to the low-level policy to generate actions. It is optimized with task-specific and motion prior rewards, directing the output skills to focus more on the downstream task. The task-specific rewards $r_G$ in reaching and grasping tasks are divided into four stages: 1) direction and walking, encouraging correct facing and target velocity; 2) pre-grasping, focusing on raising the hand to approach the grasp; 3) grasping, using a reward similar to UnidexGrasp\cite{UniDexGrasp} to incentivize successful grasping; 4) post-grasping, maintaining the grasped object with an additional reward.

In addition to task rewards, we borrow the motion prior idea from ASE\cite{ASE}, using the discriminator $D$ from low-level policy training as an extra reward $r_{p_1}$ to avoid change too frequently between skills:

\begin{equation}
    r_{p_1} = - \log (1-D(s_t,s_{t+1}))
\end{equation}
Initially, the high-level policy samples skills uniformly from the latent space for exploration, which could lead to inefficient exploration in a limited data setting due to the imperfect space. To mitigate this, we introduce a motion prior $D'$(trained alongside the task) during the first walking stage, guiding the sampling toward defined regions of continuous walking:\

\begin{equation}
    r_{p_2} = - \log (1-D'(s_t,s_{t+1}))
\end{equation}
 Specifically, the total reward is expressed as $r_t = w_{G} r_{G} + w_{p_1} r_{p_1} + w_{p_2} r_{p_2}$, where $w_{G}, w_{p_1}, w_{p_2}$ represent the reward weights for different combinations.

\subsection{Active Strategy in Providing Data}

\label{active}

Due to dataset limitations, the initial high-level policy struggles to generate natural movements for tasks beyond the walking scenario (e.g., tasks involving low or high tables). To address this, we leverage advanced kinematic pose priors to generate tailored grasping poses for these challenging scenarios, serving as useful task-specific guidance. Similar to FLEX \cite{FLEX}, we combine body and hand priors with 3D geometric constraints to generate poses across diverse environments, avoiding the need for 3D full-body grasping data, which is harder to collect and less flexible. We then apply slerp interpolation from the rest pose to the grasping pose to create continuous motion. However, these generated motions lack natural movement patterns, leading to artifacts and challenges in dynamic tasks such as mobile grasping. To overcome these limitations, we propose adding minimal generated data to the dataset, preserving its fidelity while addressing task-specific constraints. This motivates an active strategy based on performance estimation to maximize the utility of the generated motions.

We discretize the task based on key parameters, such as table height, and evaluate the success rate $sr$ and the average prediction score $\overline{p}$, specified by $\log r_{p_1}$, where $r_{p_1}$ is the reward received from the discriminator $D$ in the low-level policy. Both low success rates and low prediction scores indicate the need to augment the motion space with additional motions. These two metrics are combined in a weighted score to assess task performance.
\begin{equation}
\begin{split}
        \text{W}_j = s_0 &+ w_{succ} \frac{\max_i{sr_i} - sr_j}{\max_i{sr_i} - \min_i{sr_i}} \\
        &+ w_{disc} \frac{\max_i{\overline{p_i}} - \overline{p_j}}{\max_i{\overline{p_i}} - \min_i{\overline{p_i}}}
\end{split}
\end{equation}
where $W_j$ denotes the overall score of the $j$-th task and $s_0, w_{succ}, w_{disc}$ are constants for adjustment. $\min_i$ and  $\max_i$ denote the maximum/minimum value among all tasks. The final added data weight is proportional to $- \exp(-W_j)$, and the data with a larger weight is more likely to be sampled during training.

\subsection{Local Feature Alignment Mechanism}

\label{feats}
Through the active strategy, we extend the initial space to address more complex tasks. However, the quality of the generated motion remains unverified, lacking guarantees for physical realism and dynamic motion patterns due to reliance on static pose priors.

Observations from the pilot study show that features from the shallow layers of the critic network better capture real motion patterns. During low-level policy tuning, we introduce an additional reward to align the generated features with the pre-calculated walking feature distribution. Let the $i$-th feature have mean $\mu_i$ and variance $\sigma_i$, and let $f_i(s, z)$ extract the current feature. We compute the Mahalanobis distance to measure the deviation from the distribution:
\begin{equation}
    d^{ma}_{f_i} = \sqrt{(f_i(s, z)-\mu_i)(\sigma_i+\epsilon \mathbb{I})^{-1}(f_i(s, z)-\mu_i)}
\end{equation}

We add a constant $\epsilon$ to avoid zero-eigenvalue in $\sigma$. The reward is then:
\begin{equation}
    r^{feats} = -\sum_{f_i} w_{f_i} d^{ma}_{f_i} \mathbbm{1}(d^{ma}_{f_i}>\text{thres}_{f_i})
\end{equation}
where $w_{f}$ and $\text{thres}_{f_i}$ represent the weight for feature $f_i$ and the corresponding threshold which prevents excessive reduction in diversity. Notably, the feature extractor takes both motions and latent variable $z$ as input. For a given motion, the discriminator(only takes the state as input) only evaluates whether it matches the dataset, while the critic also assesses whether the motion is aligned with the initial space manifold. This supports low-level policy tuning, with further details provided in the supplementary materials.

\begin{table*}[t]
\adjustbox{width={\linewidth},keepaspectratio}{
\begin{tabular}{|c|cccc|cccc|}
\hline
\small
\multirow{2}{*}{Method} & \multicolumn{4}{c|}{Simple Scenes} & \multicolumn{4}{c|}{Complex Scenes} 
\\ \cline{2-9} 
& \multicolumn{1}{c|}{SR(Grasp)} & \multicolumn{1}{c|}{SR(Goal)} & \multicolumn{1}{c|}{  User(G)$\uparrow$ } & User(I)$\uparrow$ & \multicolumn{1}{c|}{SR(Grasp)} & \multicolumn{1}{c|}{SR(Goal)} & \multicolumn{1}{c|}{ User(G)$\uparrow$ } & User(I)$\uparrow$ \\ \hline
Fullbody PPO            & \multicolumn{1}{c|}{96.6\%}          & \multicolumn{1}{c|}{0.01\%}         & \multicolumn{1}{c|}{0.00\%}          &   0.25\%       & \multicolumn{1}{c|}{70.2\%}          & \multicolumn{1}{c|}{0.08\%}         & \multicolumn{1}{c|}{0.00\%}          &     0.00\%    
  \\ \hline
ASE                     & \multicolumn{1}{c|}{55.7\%}          & \multicolumn{1}{c|}{13.4\%}         & \multicolumn{1}{c|}{5.00\%}          &      0.25\%    & \multicolumn{1}{c|}{40.2\%}          & \multicolumn{1}{c|}{10.5\%}         & \multicolumn{1}{c|}{0.00\%}          &  4.50\%        \\
AMP                     & \multicolumn{1}{c|}{85.3\%}          & \multicolumn{1}{c|}{58.1\%}         & \multicolumn{1}{c|}{14.8\%}          &   16.5\%       & \multicolumn{1}{c|}{65.5\%}          & \multicolumn{1}{c|}{38.0\% }         & \multicolumn{1}{c|}{0.50\%}          &    2.00\%      \\

PMP(2-Part)                & \multicolumn{1}{c|}{63.8\%}          & \multicolumn{1}{c|}{32.2\%}         & \multicolumn{1}{c|}{ 0.25\%  }          &     0.50\%        & \multicolumn{1}{c|}{50.4\%}          & \multicolumn{1}{c|}{32.9\%}         & \multicolumn{1}{c|}{0.00\%     }          &    0.00\%           \\
PMP(5-Part)                    & \multicolumn{1}{c|}{73.6\%}          & \multicolumn{1}{c|}{45.8\%}         & \multicolumn{1}{c|}{0.25\%}          & 0.25\%  & \multicolumn{1}{c|}{68.7\%}          & \multicolumn{1}{c|}{41.0\%}         & \multicolumn{1}{c|}{0.25\%     }          &   0.00\%            \\
PSE(2-Part)                   & \multicolumn{1}{c|}{99.6\%}          & \multicolumn{1}{c|}{85.5\%}         & \multicolumn{1}{c|}{0.50\%}          &   0.00\%       & \multicolumn{1}{c|}{41.8\%}          & \multicolumn{1}{c|}{33.9\%}         & \multicolumn{1}{c|}{0.50\%     }               &     0.00\%     \\
PSE(5-Part)                    & \multicolumn{1}{c|}{99.5\%}          & \multicolumn{1}{c|}{75.7\%}         & \multicolumn{1}{c|}{0.00\%}          &    0.00\%      & \multicolumn{1}{c|}{67.2\%}          & \multicolumn{1}{c|}{42.6\%}         & \multicolumn{1}{c|}{0.00\%     }          &      0.00\%         \\
AMP* & \multicolumn{1}{c|}{85.9\%}          & \multicolumn{1}{c|}{72.3\%}         & \multicolumn{1}{c|}{10.5\%}          &  6.75\%        & \multicolumn{1}{c|}{66.7\%}          & \multicolumn{1}{c|}{55.3\%}         & \multicolumn{1}{c|}{3.00\%}          &   4.75\%       \\ \hline
Ours   & \multicolumn{1}{c|}{\textbf{99.8\%}}          & \multicolumn{1}{c|}{\textbf{88.8\%}}         & \multicolumn{1}{c|}{--}          &    --      & \multicolumn{1}{c|}{\textbf{69.7\%}}          & \multicolumn{1}{c|}{\textbf{55.8\%}}         & \multicolumn{1}{c|}{--}          &  --        \\ \hline
Oracle Grasp Policy           & \multicolumn{1}{c|}{100.0\%}          & \multicolumn{1}{c|}{95.8\%}         & \multicolumn{1}{c|}{ -- }          &   --       & \multicolumn{1}{c|}{75.8\%}          & \multicolumn{1}{c|}{72.1\%}         & \multicolumn{1}{c|}{--}          &  -- \\     
Oracle Policy          & \multicolumn{1}{c|}{97.4\%}          & \multicolumn{1}{c|}{59.0\%}         & \multicolumn{1}{c|}{38.0\%}          &  55.2\%        & \multicolumn{1}{c|}{69.7\%}          & \multicolumn{1}{c|}{53.4\%}         & \multicolumn{1}{c|}{52.5\%}          &    55.5\%      \\ \hline
\end{tabular}}
\caption{\textbf{Quantitative results of overall task performance:} We compare success rate and user preference. Our method(with $f_0 \& f_1$ aligned, data ratio 20\%) achieves the highest success rate while remaining significantly more naturalness among all baselines.}

\label{tab:overall_result}
\end{table*}

\section{Experiments}

\subsection{Settings}

\textbf{Dataset:} Our walking dataset contains several walking segments (about 30s) from the AMASS~\cite{AMASS} and the CMU MoCap dataset~\cite{CMU}, including walking along with turns. For pose generation, similar to FLEX~\cite{FLEX}, we use VPoser~\cite{pavlakos2019expressivebodycapture3d} as body prior and GrabNet~\cite{GRAB} as hand prior respectively.


\noindent \textbf{Character:} Our framework uses a 3D humanoid character with 15 joints, similar to~\cite{AMP}, but replaces the spherical hand with a dexterous hand~\cite{shadowrobot2005} featuring 23 joints. The spherical joint connecting the forearm to the hand has two degrees of freedom (DoF).

\noindent\textbf{Training:} The first iteration of low-level policy training uses the dataset $M_0$ and trains in \textbf{simple scenes}, where the agent walks from the starting point to a 0.8 m wide table, grasps a specific object on tables with height varing from 0.05 m to 1.65 m, and returns. We then actively generated data and added them with a various data ratios (added motion weighted length/initial motion weighted length). After data augmentation, the next iteration of task training was conducted simultaneously in both the \textbf{simple scenes} (for testing the effects in compensating for the limitations of the initial space) and the \textbf{complex scenes} (for testing generalizability). In the complex scenes, the table size randomly ranged from 0.6 m to 1.2 m, and the start and end points were positioned around the table at any angle between 1.5 m and 4 m. We follow UnidexGrasp~\cite{UniDexGrasp} and use a 3-object curriculum learning approach to train on a similar object dataset (5519 objects across 133 categories) and test on objects from unseen categories.

\subsubsection{Metrics}

To measure task completion, we use two metrics: \textbf{SR(Grasp)}, which calculates the percentage of episodes in which the agent successfully grasps and lifts the object by at least 0.1 m, and \textbf{SR(Goal)}, which measures the percentage of episodes where the agent reaches within 0.5 m of the goal while maintaining balance until the episode ends.

We also need to evaluate the motion quality of reaching and grasping. The distribution similarity metrics (such as FID) and common inception models, like those in~\cite{chen2023executing, tevet2022humanmotiondiffusionmodel}, tend to overfit to their training data and fail to properly evaluate our method. Following prior works~\cite{aberman2020unpaired, hong2022avatarclip, tevet2022motionclip}, we conducted a user study with 100 volunteers, who compared our generated motions with a baseline side-by-side and selected the better one based on two criteria: 1) General Motion Quality (\textbf{User(G)}), which evaluates the overall motion quality, including aspects like grasp, walking forward, and walking back; 2) Motion Issues (\textbf{User(I)}), which assesses if the motion sequence shows evident unnaturalness, including aspects like shuffling, stumbling, or balance problems. Specifically, $\text{User(G or I)} = \sum_{i} w_i\cdot \text{score}_i$,
where $w_i$ is the weight of the \( i \)-th aspect and \( \text{score}_i \) is the percentage of users selecting baseline as the better one for the \( i \)-th aspect.


\subsubsection{Baselines}

We first compared our method with the physics-based motion generation methods trained on the same dataset with the same reward design. We also included pure RL (\textbf{Fullbody PPO}) and Oracle policies as references.

\noindent \textbf{AMP}~\cite{AMP}, \textbf{ASE}~\cite{ASE}: These methods are capable of reproducing combinations of skills based on tasks and reference data. For a fair comparison, we augment AMP with generated data (matching our policy), denoted as \textbf{AMP*}.

\noindent \textbf{PMP}~\cite{pmp}, \textbf{PSE}: PMP generates more flexible motions using part-wise motion priors and dynamic supervision. We also introduce part-wise discriminators into ASE’s skill-embedding framework, forming PSE. We utilize 2-part (Upper \& Lower) and 5-part configurations(shown in Figure~\ref{fig:Critic-architecture}).

  We follow UnidexGrasp~\cite{UniDexGrasp} to train a non-goal-conditioned grasp policy (\textbf{Oracle Grasp Policy}) and to further investigate the gap between real and generated data, we also add real MoCap CIRCLE ~\cite{circle} to augment the walking dataset for policy training(\textbf{Oracle Policy}).

\subsection{Overall Results and Analysis}

\begin{figure*}[h!t]
    \centering
    \includegraphics[width=0.8\linewidth]{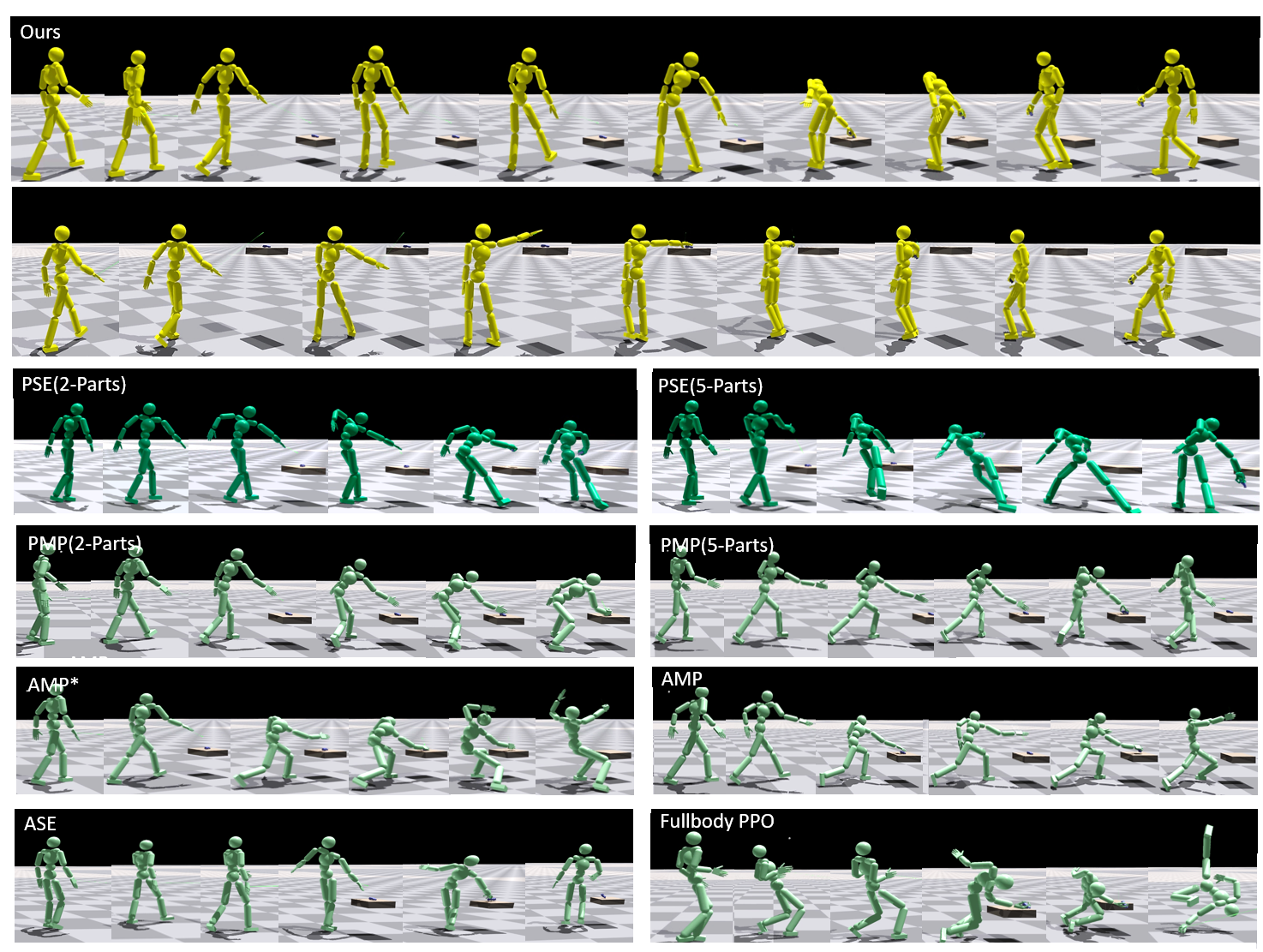}
    \caption{\textbf{Visualization of the overall task compared to baselines}: We visualized our method and baselines. Our methods can yield nature reaching and grasping in various scenes and tasks while baselines show significant unnatural movements.}

    \label{fig:tasks}
\end{figure*}

We evaluated the baselines using the above metrics, with quantitative results presented in Table~\ref{tab:overall_result} and visualization in Figure~\ref{fig:tasks}. More videos and comparative results are available in the supplementary materials.

We observed that \textbf{Fullbody PPO} struggles with exaggerated motions and lowest SR(Goal) due to the lack of demonstration guidance. \textbf{ASE} experiences significant divergence between walking motion reproduction and task completion, resulting in a low success rate. \textbf{AMP} faces a similar issue but performs better in success rate for being less constrained by its discriminator, which trains alongside the task. In \textbf{PMP} and \textbf{PSE}, part-wise discriminators provide more flexible supervision. However, as the number of parts increases, selecting the right part-wise motion to maintain balance becomes more challenging, leading to unnatural movements or balance issues, which limits it success rate, especially in PMP. To compare learning through a motion prior versus a motion space, we evaluated \textbf{AMP*}. The augmented dataset is highly diverse and complex, containing artifacts that make it challenging for AMP to learn the motion prior alongside the task, resulting in unnatural movements. Additionally, AMP always trains from scratch and re-learns walking skills for each task, limiting its efficiency.


We also compared our method with \textbf{Oracle Grasp Policy}. Despite the challenge of balancing, we approximate the Oracle Policy with high grasp success rates. We also found that in simple scenes, our task-specific generated data outperforms MoCap data, highlighting the limitations of real data with fixed distributions. Even in more complex scenes, our success rate is comparable to or even exceeds that of the \textbf{Oracle Policy}. This may be due to the feature alignment mechanism (Section~\ref{sec:feats}) which transfers balance skills from walking to support reaching and grasping tasks.

\subsection{Compare to concurrent SOTA work}
WANDR~\cite{Diomataris_2024_CVPR}, Braun et al.~\cite{phy}, and Omnigrasp~\cite{omnigrasp} share similar settings with our work.
We directly adopt WANDR's reaching results (kinematic formulation) for comparison, while Braun et al., Omnigrasp, and our method are physics-based.
For Braun et al., we re-implemented their approach using our dataset as a body prior and evaluated it on high/low tables (beyond GRAB's original scope).
For Omnigrasp~\cite{omnigrasp}, we leveraged its inference model for close-range grasping due to code unavailability.
Quantitative results (Table~\ref{tab:comparison}) demonstrate that our framework achieves: (Top 3: \colorbox{best}{Red} $>$  \colorbox{second}{Orange} $>$  \colorbox{third}{Yellow}.)
Superior naturalness validated by user studies and GPT-4o\cite{openai2025gpt4o}/Kimi \cite{moonshot2025kimi} scores (0-10 scale, higher is better);
Fewer artifacts and more realistic motion transitions compared to Braun et al. and Omnigrasp;
Balanced performance across \textbf{naturalness, physical plausibility, and success rates}.

\begin{table}[ht]
  \centering
  \adjustbox{width={\linewidth},keepaspectratio}{
    \begin{tabular}{|c|c|c|c|c|c|}
    \hline
        \multicolumn{1}{|c|}{Methods} & \multicolumn{1}{c|}{\textbf{SR(Grasp) $\uparrow$}}  & \multicolumn{1}{c|}{\textbf{SR(Goal)$\uparrow$}}  & \multicolumn{1}{c|}{\textbf{FS($\pm$1\%) $\downarrow$}} 
       & \multicolumn{1}{c|}{\textbf{GPT-4o/Kimi $\uparrow$}} 
       & \multicolumn{1}{c|}{\textbf{User $\uparrow$}}\\
        \hline
        \textbf{Ours} & \cellcolor{best}{69.7}\%    & \cellcolor{best}{55.8}\%    &\cellcolor{second}{12.0\%}   &\cellcolor{second}{7.38/7.25}    &\cellcolor{third}{7.55}\\
        \textbf{Oracle Policy}      & \cellcolor{best}{69.7}\%    & \cellcolor{second}{53.4}\%     & \cellcolor{best}{11.5\%}   &\cellcolor{third}{7.50/7.00}    &\cellcolor{second}{7.75}\\
        \textbf{WANDR}      &{32\%(reach)}    & {--}    &{16\%(reach)}  & \cellcolor{best}{8.03/7.75} & \cellcolor{best}{8.33}\\
        \textbf{Braun et al.}   & \cellcolor{third}{59.6\%}    &{22.2\%}     &{14.5\%}   &{6.00/5.00}    &  {5.83}\\
        \textbf{Omnigrasp}  &   {54.4}\%     & \cellcolor{third}{52.6\%}   &\cellcolor{third}{13.2\%}  & {6.50/6.13}  & {5.67}\\
    \hline
    \end{tabular}}

\caption{\textbf{Results of comparison with concurrent SOTAs}}
  
\label{tab:comparison}
\end{table}


\subsection{Ablation Study}

\subsubsection{Active Augmentation}

To validate the effectiveness of our active design, we compare the success rate in simple scenes across different data-adding strategies. This comparison highlights the strategy's impact on the success rate and evaluates the effectiveness of our active design, as shown in Table~\ref{abl:active}.
\begin{table}[!ht]
\centering
\small
\begin{tabular}{|cc|c|c|}
\hline
\multicolumn{2}{|c|}{Strategies/Ratio(\%)}                 & SR(Grasp) & SR(Goal)  \\ \hline
\multicolumn{1}{|c|}{\multirow{3}{*}{Random}}      & 5  & 55.6\%      & 15.3\%     \\ \cline{2-4} 
\multicolumn{1}{|c|}{}                             & 10 & 81.2\%      & 20.7\%        \\ \cline{2-4} 
\multicolumn{1}{|c|}{}                             & 20 & 92.1\%      & 64.1\%      \\ \hline
\multicolumn{1}{|c|}{\multirow{3}{*}{Active-S}}    & 5  & 70.1\%      &   30.1\%      \\ \cline{2-4} 
\multicolumn{1}{|c|}{}                             & 10 & 92.8\%      & 36.6\%        \\ \cline{2-4} 
\multicolumn{1}{|c|}{}                             & 20 & \textbf{95.2\%}      & 64.2\%      \\ \hline
\multicolumn{1}{|c|}{\multirow{3}{*}{Active-D}}    & 5  & 57.9\%      &   32.0\%      \\ \cline{2-4} 
\multicolumn{1}{|c|}{}                             & 10 & 87.1\%      & 38.5\%       \\ \cline{2-4} 
\multicolumn{1}{|c|}{}                             & 20 & 90.2\%      & 66.6\%       \\ \hline
\multicolumn{1}{|c|}{\multirow{3}{*}{Active-Both}} & 5  & \textbf{69.9\%}      & \textbf{36.3\%}      \\ \cline{2-4} 
\multicolumn{1}{|c|}{}                             & 10 & \textbf{95.2\%}      &\textbf{39.3\%}     \\ \cline{2-4} 
\multicolumn{1}{|c|}{}                             & 20 & 92.4\%      & \textbf{69.1\%}   \\ \hline
\end{tabular}

\caption{\textbf{Results of active strategy ablation:} We compare our method, \textbf{Active-Both} combining both success rate and discriminator score to actively adding data, with three strategies: random(\textbf{Random}), solely based on success rate (\textbf{Active-S}), and solely on discriminator score (\textbf{Active-D}). The active strategy based on both yields higher task success rates.}
  
\label{abl:active}
\end{table}
We found that the active strategy improves the success rate with the same amount of generated data, especially when data is scarce. Both success estimation and discriminator scores are valuable, as success rates can be limited by space constraints or discriminator scores will reflect the motions beyond space manifold. Therefore, a comprehensive approach considering these factors is key to effective data augmentation. We also note that increasing the added data volume leads to longer training times before convergence, and higher data ratio may even degrade performance, as further detailed in the supplementary materials.

\subsubsection{Feature Alignment Mechanism}

\label{sec:feats}
\begin{figure}
    \centering
    \small
    \includegraphics[width=0.8\linewidth]{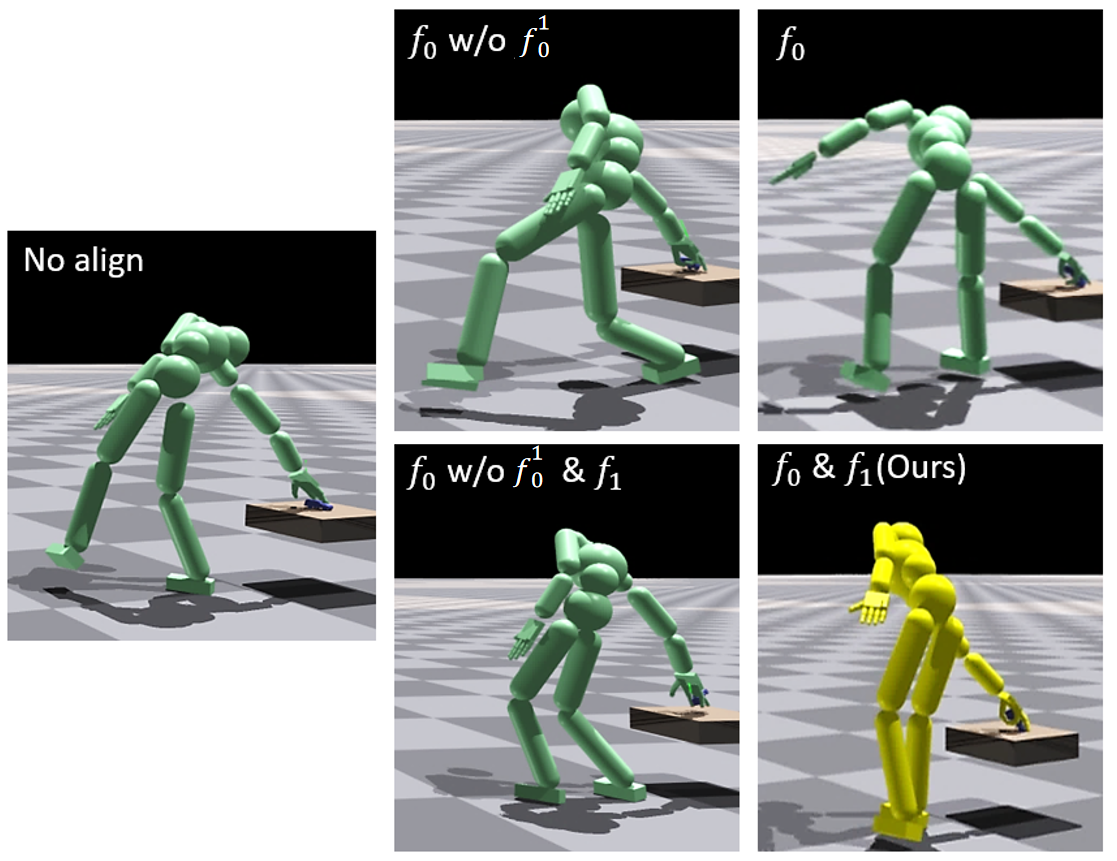}
    \caption{\textbf{Visualization of effects of feature align mechanism.}}
    \label{fig:feats}
   
\end{figure}

To validate the effectiveness of the feature alignment mechanism, we compared the results across various choices of aligned features. The definitions of $f_i$ are shown in Figure~\ref{fig:Critic-architecture}. We evaluated the success rate and designed a detailed user study, that first asked part of volunteers to identify the five most unnatural features by watching demos of interpolated and real data and then asked the others to rank six motion segments according to this, scoring them from 0 to 5, with 5 for the best and 0 for the worst (\textbf{User(I)}). Additionally, we trained a classifier using a pre-trained feature extractor from MotionGPT~\cite{motiongpt} on the CIRCLE~\cite{circle} real dataset (positive examples) and slerp-interpolated CIRCLE data (negative examples). This classifier was used to score our motion data, revealing that the main artifacts stem from the interpolation (\textbf{Pred}). The results are shown in Table~\ref{tab:feats}) and visualization in Figure~\ref{fig:feats} are comparisons for the same grasp pose of 20\% data ratio for different choices.

We observed that incorporating $f_0$ without $f_0^1$ alignment reduces artifacts in generated data, such as a stiff left arm in static poses, as confirmed by our pilot study. Adding the torso feature ($f_0^1$) improves handling of over-bent torsos in mobile grasping, enhancing stability and boosting success rates. Additionally, aligning $f_1$ ensures natural left arm extension for balance, with the left foot grounded, preventing excessive lifting in static poses. We conclude that this alignment mechanism transfers knowledge from walking motions to improve reaching and grasping. Static grasping poses require precise full-body coordination, which is difficult to learn with limited references, especially after transitioning from walking. Walking, as a core locomotion skill, provides more dynamic balance information, making it more effective than part-wise supervision alone.

\begin{table}[h!t]
\small
\small
\centering
\begin{tabular}{|cc|c|c|c|}
\hline
\multicolumn{2}{|c|}{Aligned feats/Ratio(\%)}                 & SR(Goal) &User(I)$\uparrow$ & Pred$\uparrow$ \\ \hline
\multicolumn{1}{|c|}{\multirow{3}{*}{No align}}      & 5      & 36.3\% & 1.16    &  0.8549   \\ \cline{2-5} 
\multicolumn{1}{|c|}{}                             & 10      & 39.3\%  & 1.20   &  0.8039   \\ \cline{2-5} 
\multicolumn{1}{|c|}{}                             & 20      & 69.1\% & 1.17    &  0.7641   \\ \hline
\multicolumn{1}{|c|}{\multirow{3}{*}{$f_0$ w/o $f^1_{0}$}}      & 5     & 43.3\%   & 1.85   &  0.8534   \\ \cline{2-5} 
\multicolumn{1}{|c|}{}                             & 10     & 51.0\%  & 1.79  & 0.8216    \\ \cline{2-5} 
\multicolumn{1}{|c|}{}                             & 20       & 79.8\%  &1.83  & 0.8370    \\ \hline
\multicolumn{1}{|c|}{\multirow{3}{*}{$f_0$}}    & 5      & 37.2\%  & 3.52   &  0.8572   \\ \cline{2-5} 
\multicolumn{1}{|c|}{}                             & 10      & 51.8\% & 3.50   &   0.8532  \\ \cline{2-5} 
\multicolumn{1}{|c|}{}                             & 20       & 84.6\% & 3.60    & 0.8759\\ \hline
\multicolumn{1}{|c|}{\multirow{3}{*}{$f_0$ w/o $f^1_{0}$ \&$f_1$}}    & 5       & 40.1\% & 3.75 & 0.8949     \\ \cline{2-5} 
\multicolumn{1}{|c|}{}                             & 10   & 50.0\% & 3.73    & 0.8871  \\ \cline{2-5} 
\multicolumn{1}{|c|}{}                             & 20        &  74.2\% & 3.46 & 0.8859        \\ \hline
\multicolumn{1}{|c|}{\multirow{3}{*}{$f_0$ \& $f_1$}} & 5  &   63.6\%  & 4.72   & 0.9063   \\ \cline{2-5} 
\multicolumn{1}{|c|}{}                             & 10   & 82.8\% & 4.77   &  0.8820\\ \cline{2-5} 
\multicolumn{1}{|c|}{}                             & 20      & \textbf{88.8\%} & \textbf{4.94}  & 0.8488    \\ \hline
\multicolumn{1}{|c|}{\multirow{3}{*}{$f_0$, $f_1$ \& $f_2$}} & 5     & 34.0\%   & 0.00   &  0.7631  \\ \cline{2-5} 
\multicolumn{1}{|c|}{}                             & 10    & 38.1\%  & 0.01   & 0.5463  \\ \cline{2-5} 
\multicolumn{1}{|c|}{}                             & 20    &  47.5\%  & 0.00   &  0.5021   \\ \hline

\end{tabular}

\caption{\textbf{Quantitative results of the feature alignment ablation:} The mechanism improves success rate and reduces artifacts.}

\label{tab:feats}
\end{table}

%

\section{Conclusion}
\vspace{-2mm}
In this work, we present a framework that generates diverse, physically feasible full-body human reaching and grasping motions using only brief walking MoCap data. By leveraging active-generated data, we overcome dataset limitations, enabling reaching and grasping while learning natural motion patterns from walking. Our results show that this approach captures valuable information, producing high-fidelity motion  with extremely limited data, which can benefit humanoid robot learning. This work also opens avenues for future research, such as using more complex network designs (e.g., ST-GCN) to extract more precise patterns.


{
    \small
    \bibliographystyle{ieeenat_fullname}
    \bibliography{main}
}

\appendix
\clearpage
\maketitlesupplementary
\section{Methods Details}
\label{secA}
\subsection{Character and Degrees of Freedom}
To assess the effectiveness of our approach to modeling a character's walking and manipulation behaviors, we use our framework to train sophisticated 3D simulated characters to perform a range of tasks. 
This involves using a character based on AMP~\cite{AMP}, where we have modified the spherical hand structure into a more dexterous configuration, such as the Shadow Hand~\cite{shadowrobot2005}. 
The humanoid model consists of a body with 15 joints featuring 28 degrees of freedom (DoF) and a dexterous hand with 23 joints providing 25 DoF.

\subsection{State and Observation}
As our humanoid character is an adaptation of AMP~\cite{AMP} and ShadowHand~\cite{shadowrobot2005}, the state is primarily derived from AMP body states with modifications to the right hand using UniDexGrasp hand states~\cite{UniDexGrasp}. 
This state comprises a collection of features detailing the configuration of the character's body and its dexterous hand. These features encompass:
\begin{itemize}
  \item {Height of the root from the ground.}
  \item {Rotation of the root in local coordinates.}
  \item {Root's velocity (linear and angular) in local coordinates.}
  \item {Local rotation of each joint.}
  \item {Local velocity of each joint.}
  \item {Local rotation of right hand joints.}
  \item {Local velocity of right hand joints.}
  \item {Local position of right hand joints.}
  \item {Right hand joint DoF.}
  \item {Velocity of right hand joint DoF.}
  \item {Positions of feet, left hand, and right fingertips in local coordinates.}
\end{itemize}

In low-level policy training, observations consist of body states. 
For high-level downstream tasks, observations encompass full-body states along with task-specific data. 
For instance, in a grasping task, observations may include the object's and table's position and rotation, as well as the distance between the right fingertips and the target object.

\subsection{Stage-based Task-specific Reward}

As outlined in the primary document, the high-level policy is optimized using rewards tailored to specific tasks, along with motion-prior rewards. 
In reaching and grasping tasks, task-specific rewards $r_G$ are divided into four distinct phases:

\begin{enumerate}
    \item direction and walking, encouraging correct facing and target velocity.
    \item pre-grasping, focusing on moving the hand to approach the grasp pose.
    \item  grasping, employing a reward mechanism similar to UnidexGrasp\cite{UniDexGrasp} to promote effective grasping.
    \item  post-grasping, aiming at maintaining the grasped object, with an additional reward for preserving body equilibrium.
\end{enumerate}

In the first stage, the location reward is designed to guide the agent toward the target by combining three components: the position reward, the velocity reward, and the facing reward. The combined reward is formulated as:
\begin{equation}
    r_\text{{location}} = w_{\text{pos}} \cdot r_{\text{pos}} + w_{\text{vel}} \cdot r_{\text{vel}} + w_{\text{face}} \cdot r_{\text{face}},
\end{equation}

where:
\begin{equation}
    r_{\text{pos}} = \exp\left(0.5 \times  \|\Tilde{p}_{\text{tar}} - \Tilde{p}_{\text{root}}\|^2\right),
\end{equation}
\begin{equation}
    r_{\text{vel}} = \exp\left(-4.0\times (v_{\text{{tar}}} - (n_{\text{tar}} \cdot \Tilde{v}_{\text{root}}))^2\right),
\end{equation}
\begin{equation}
    r_{\text{face}} = 1 + (n_{\text{tar}} \cdot  \Tilde{f}).
\end{equation}

Here, $v_{\text{tar}}$ is the target velocity. $\Tilde{v}_{\text{root}}$, $\Tilde{p}_{\text{tar}}$ and $\Tilde{p}_{\text{root}}$ represent the x-y plane velocity of the root, the x-y plane positions of the target and the root, respectively. $\Tilde{f}$ represents the projection of the facing direction onto the x-y plane. \(n_{\text{tar}}\) is the normalized direction to the target:
\begin{equation}
    n_{\text{tar}} = \frac{\Tilde{p}_{\text{tar}} - \Tilde{p}_{\text{root}}}{\|\Tilde{p}_{\text{tar}} - \Tilde{p}_{\text{root}}\|}.
\end{equation}

Additional masks are applied to adjust the rewards based on proximity or alignment:
- If \(\|\Tilde{p}_{\text{tar}} - \Tilde{p}_{\text{root}}\| < 0.5\), all rewards are boosted.
- If the agent is moving in the wrong direction (\(n_{\text{tar}} \cdot \Tilde{v}_{\text{root}}) \leq 0\)), the velocity reward is set to zero.

In the second stage, the reward can be computed by $r_{\text{reach}} = \exp\left( - \|{p}_{\text{tar}} - {p}_{\text{reach}}\|^2\right)$. Here, $p_{\text{reach}}$ represents the position of the reaching end-effector (the palm center), while ${p}_{\text{tar}}$ represents the ideal reaching position. The ideal position is defined as 0.2 m above the object, at a distance of one-third of the table's width from its center, and oriented towards the direction from which the person approached.

In the third stage, the reward is defined by grasping-related metrics, similar in UnidexGrasp~\cite{UniDexGrasp}. This reward, \(r_{\text{grasp}}\), evaluates the agent's ability to successfully lift and stabilize an object while maintaining effective grasp control. It combines three components: the grasp quality reward, the object height reward, and the object velocity penalty. These components are weighted and combined as follows:
\begin{equation}
    r_{\text{grasp}} = w_{\text{grasp}} \cdot r_{\text{grasp\_quality}} + w_{\text{height}} \cdot r_{\text{height}} + w_{\text{obj-vel}} \cdot r_{\text{obj-vel}},
\end{equation}

where:
\begin{equation}
    r_{\text{grasp\_quality}} =2 -0.5 \cdot d_{\text{finger}} - 1.0 \cdot d_{\text{hand}},
\end{equation}
\begin{equation}
    r_{\text{height}} = 0.1 + 0.5 \cdot \frac{h_{\text{object}}}{h_{\text{lift\_target}}},
\end{equation}
\begin{equation}
    r_{\text{obj-vel}} = -0.2 \cdot \text{clamp}(v_{\text{object}} - v_{\text{threshold}}, 0, 5).
\end{equation}

Here, \(d_{\text{finger}}\) denotes the distance between the agent's fingers and the object, while \(d_{\text{hand}}\) represents the distance between the agent's hand and the object. \(h_{\text{object}}\) is the height of the object above the table, normalized by the target lifting height \(h_{\text{lift\_target}}\). The object height reward, \(r_{\text{height}}\), is applied only if the agent is in contact with the object. \(v_{\text{object}}\) indicates the velocity of the object and \(v_{\text{threshold}}\) is a velocity threshold used to penalize excessive movement. This reward structure effectively balances the need for precise grasping, stable lifting, and minimal object movement, guiding the agent to complete the task efficiently and effectively.

In the final stage, the reward is defined as \(r_{\text{goal}}\), which combines the proximity of the agent to the target and its grasp quality. The reward is computed as:

\begin{equation}
    r_{\text{goal}} = 3 \cdot r_{\text{location}} + 3 \cdot \text{clamp}(1.5 + r_{\text{grasp\_quality}}, 0, 5).
\end{equation}
Here, $r_{\text{location}}$ represents the location reward as defined in the first stage, with the target position updated to match the goal in this stage.

Furthermore, we include an additional reward, $r_{\text{stage}}$, defined as a bonus granted when the transition condition is satisfied.
\subsection{Transition Condition}
To ensure smooth progression between stages, a set of transition conditions is defined. These conditions evaluate whether the agent has successfully completed the current stage's objectives and is ready to move to the next stage. The conditions are:
\begin{itemize}
    \item The distance between the root and the object is less than $1$ meter.
    \item The palm is directly above the object, with a vertical distance of less than 0.1 meters.
    \item The object is lifted vertically by more than 0.1 meters relative to its initial position.
\end{itemize}
\subsection{Feature Alignment Mechanism}

We aim for the generated motions to accomplish diverse tasks at a macro-level, similar to the generated data, while preserving realistic patterns at a micro-level. These patterns represent common traits observed in real-world data. For instance, humans naturally synchronize their right hand and left foot to maintain balance, move their forearm driven by the upper arm, and exhibit joints that rarely bend or have limited bending angles. Intuitively, we hypothesize that such patterns arise from local observations. To incorporate this understanding, we modified the architecture of our critic network.

Previously, the network's first layer directly processed the entire 223-dimensional full-body observation through three layers of MLPs. To better capture the hierarchical nature of motion patterns, we introduce five separate sub-networks, each dedicated to processing the local observations of specific body parts (torso and four limbs). Each sub-network performs an initial transformation on its respective input, producing part-specific features $f_0$. These features are then passed into the subsequent shared layers of the network, enabling a more structured and progressive analysis.

This revised architecture allows the network to focus first on part-wise information, capturing localized patterns. Then it gradually integrates information across different parts, enabling the model to form increasingly global representations. This hierarchical progression ensures that both local patterns and global coherence are effectively captured.

As indicated in the pilot study, MoCap-Reach and MoCap-Walk exhibit evident clustering in the shallow layers of the network, but this clustering diminishes in the deeper layers. This observation suggests that, despite differences in orientation, real motions share common patterns, particularly at the part-wise level. Additionally, the first level of the network also demonstrates some degree of clustering, which provides insights into shared coordinate systems across different body parts. These patterns are referred to as "local features."

Inspired by this clustering behavior, we introduce a regularization term that encourages the generated motion to exhibit similar local feature locations. Specifically, we aim for the distance between two sets of features to remain within the variance of walking data. During space tuning, we derive the $10$-step motion and input it into the original critic, which is trained on walking motions, to calculate the mean value of the features as the current feature $f_i(s, z)$.

We then introduce an additional reward to align the generated features with the pre-calculated walking feature distribution. Let the $i$-th feature have mean $\mu_i$ and variance $\sigma_i$ respectively. The current feature is extracted, and the Mahalanobis distance is used to quantify its deviation from the feature distribution:

\begin{equation}
    d^{ma}_{f_i} = \sqrt{(f_i(s, z)-\mu_i)(\sigma_i+\epsilon \mathbb{I})^{-1}(f_i(s, z)-\mu_i)}
\end{equation}

We add a constant $\epsilon$ to avoid zero eigenvalues in $\sigma$. The reward is then:

\begin{equation}
    r^{feats} = -\sum_{f_i} w_{f_i} d^{ma}_{f_i} \mathbbm{1}(d^{ma}_{f_i}\geq\text{thres}_{f_i})
\end{equation}

where $w_{f_i}$ and $\text{thres}_{f_i}$ represent the weight for feature $f_i$ and the corresponding threshold, which prevents excessive reduction in diversity. Notably, the feature extractor takes both motions and latent variable $z$ as input. For a given motion, the discriminator (which only takes the state as input) evaluates whether it matches the dataset, while the critic also assesses whether the motion is aligned with the initial space manifold. Since we can pre-compute the inverse of the covariance matrix, this avoids the need to repeatedly compute the inverse at each step, significantly saving computation time. In our approach, $w_{f_i}$ is adjustable during training. Our best-performing implementation aligns the features only in the first two layers.

It is worth noting that many studies improve motion flexibility by leveraging local motion information. For instance, Jang et al.'s Motion Puzzle \cite{motionpuzzle} and Lee et al.'s physics-based controllers \cite{lee2022learning} focus on enhancing motion adaptability. PMP \cite{pmp}, on the other hand, provides greater flexibility for non-repetitive motions. However, these methods often overlook the interdependent patterns between body parts, leading to challenges in maintaining balance and producing natural motion.

\subsection{Data Generation}

Our method is similar to FLEX~\cite{FLEX}. Utilizing pre-trained hand-grasping~\cite{pavlakos2019expressivebodycapture3d} and human pose priors~\cite{GRAB}, our approach employs a gradient-based optimization process across multiple objectives to minimize losses related to hand-object interaction, balance constraints, and task alignment to synthesis a grasping pose.

Once a grasping pose is synthesized, we interpolate it into continuous motions. Specifically, denote the target pose with root translation $p^{\text{tar}}_{\text{root}} = (x_{\text{root}}, y_{\text{root}}, z_{\text{root}})$ and joint local rotations $q^{\text{tar}}_{i}$. Using spherical linear interpolation (SLERP), we generate \(T\) frames of motion starting from an initial standing position with root translation \(p^{\text{init}}_{\text{root}} = (0, 0, 0)\) and joint local rotations \(q^{\text{init}}_i\).
At frame $t$, the root translation $p^{t}_{\text{root}} = (x^{t}_{\text{root}}, y^{t}_{\text{root}}, z^{t}_{\text{root}})$ and joint local rotations $q^{t}_{i}$ are computed as follows:

\begin{equation}
    p^{t}_{\text{root}} = \text{slerp}(p^{\text{init}}_{\text{root}}, p^{\text{tar}}_{\text{root}}, t/T)
\end{equation}

\begin{equation}
    q^{t}_{i} = \text{slerp}(q^{\text{init}}_i, q^{\text{tar}}_i, t/T)
\end{equation}

where $\text{slerp}(\cdot, \cdot, \alpha)$ denotes the interpolation and $t$ is the current frame in the interpolation from the initial position to the target pose over $T$ frames.

Following this, we retarget the SMPL-X parameters to our humanoid model, similar to InterScene~\cite{Interscene}, ensuring that the generated motions align with the desired body structure and task requirements. To ensure physical plausibility, we enforce constraints on the generated motion: the target pose is set to rest on the left hand, and the minimum height of both feet remains consistent with the ground throughout the motion. By adhering to these constraints, the generated motion not only respects physical limitations but also aligns closely with task-specific objectives, such as precise hand-object interactions.

\section{Implementation Details}
\label{secB}
\subsection{Dataset}
Table~\ref{tab:dataset} presents the motion capture files used in our dataset. The dataset emphasize straightforward walk motion together with various but simple turning actions, constructing the brief walk reference.

Each motion sequence lasts approximately 2-5 seconds, capturing detailed and nuanced human locomotion dynamics. The inclusion of varied turning motions alongside straight walking ensures a well-rounded dataset suitable for walking to, reaching for, grasping, turning and walking back. The weights assigned to each motion type reflect their relative importance or frequency, with larger weights assigned to simple walking sequences and smaller weights to specific turning actions. This approach provides a balanced representation, aiding in effective training and evaluation.
\begin{table*}[ht]
\centering
\small
\begin{tabular}{ll}
\toprule
\textbf{File Name} & \textbf{Weight} \\
\midrule
ACCAD\_Female1Walking\_c3d\_B9\_-\_walk\_turn\_left\_(90) & 0.01463157 \\
ACCAD\_Female1Walking\_c3d\_B10\_-\_walk\_turn\_left\_(45) & 0.01463157 \\
ACCAD\_Female1Walking\_c3d\_B11\_-\_walk\_turn\_left\_(135) & 0.01463157 \\
ACCAD\_Female1Walking\_c3d\_B12\_-\_walk\_turn\_right\_(90) & 0.01463157 \\
ACCAD\_Female1Walking\_c3d\_B13\_-\_walk\_turn\_right\_(45) & 0.01463157 \\
ACCAD\_Female1Walking\_c3d\_B14\_-\_walk\_turn\_right\_(135) & 0.01463157 \\
ACCAD\_Female1Walking\_c3d\_B15\_-\_walk\_turn\_around\_(same\_direction)\_s1 & 0.02663157 \\
ACCAD\_Female1Walking\_c3d\_B15\_-\_walk\_turn\_around\_(same\_direction)\_s2 & 0.02663157 \\
ACCAD\_Female1Walking\_c3d\_B3\_-\_walk1 & 0.05263157 \\

ACCAD\_s007\_QkWalk1 & 0.05263157 \\
amp\_humanoid\_walk & 0.05263157 \\
CMU\_07\_01 & 0.10263157 \\
CMU\_07\_02 & 0.10263157 \\
CMU\_07\_07 & 0.10263157 \\
\bottomrule
\end{tabular}
\caption{\textbf{Walking Dataset:} Weights for different walking and turning.}
\label{tab:dataset}
\end{table*}

\subsection{Network Architecture}

The network architecture builds upon the design used in ASE~\cite{ASE}, with modifications tailored to meet the requirements of our system.

The \textbf{low-level policy} is implemented as an actor network that maps a state \(s\) and latent \(z\) to a Gaussian distribution over actions. This policy is realized using a fully connected network with three hidden layers of sizes [1024,1024,512] (the same configuration as the encoder and high-level policy), followed by linear output units. The \textbf{critic} for the value function divides its input into five components. Each component is processed independently through a small fully connected network. The resulting outputs are concatenated and passed through a fully connected layer with a single linear output unit, providing the value.

The \textbf{encoder} \(q(z|s, s')\) and \textbf{discriminator} \(D(s, s')\) are jointly modeled by a shared network. Separate outputs are used to compute the encoder’s mean \(\mu_q(s, s')\), normalized to \(\|\mu_q(s, s')\| = 1\), and the discriminator’s sigmoid output.

The \textbf{high-level policy} uses two hidden layers of sizes \([1024, 512]\) to generate unnormalized latents \(\bar{z}\). These are normalized to \(z = \bar{z}/\|\bar{z}\|\) before being passed to the low-level policy.

\subsection{Simulation Environment}
The experiments utilize Isaac Gym~\cite{makoviychuk2021isaac}, which is a highly efficient physics simulator that operates on a GPU. 
The training process incorporates 4096 simultaneous environments executed on one NVIDIA V100 GPU, achieving a simulation rate of $120Hz$. Every neural network is developed using PyTorch \cite{paszke2019pytorch}.

\subsection{Training Details and Hyper-Parameters}
\subsubsection{Initial Space Training}
We trained the initial space (the initial low-level policy) using a walking dataset. The training process spanned 10,000 epochs and took approximately 12 hours. Details of the hyper-parameters are provided in Table~\ref{tab:lowlevel}.
\begin{table}[ht]
\centering
\begin{tabular}{ll}
\toprule
\textbf{Hyper-parameters} & \textbf{Value} \\
\midrule
 {Learning Rate} & 2e-5 \\
Episode Length & 300\\
Action Distribution Variance & 0.055\\ 
 {Discount $\gamma$} & 0.99 \\

 {TD($\lambda$)} & 0.95 \\

Disc/Enc Mini-batchsize & 4096\\
Policy Mini-batchsize & 16384\\
 {Disc Grad Penalty Weight} & 5 \\

 {Latent Dimension} & 64 \\

 {Diversity Objective Bonus} & 0.01 \\
 {Disc Weight Decay} & 0.0001 \\
{Enc Weight Decay} & 0.000 \\
 {Disc Reward Weight} & 0.5 \\
 {Enc Reward Weight} & 0.5 \\
\bottomrule
\end{tabular}
\caption{Hyper-parameters for Low-level Policy Training}
\label{tab:lowlevel}
\end{table}
\subsubsection{Tuning with feature alignment}

We tuned the initial space on the augmented dataset. This training uses 3000-6000 epochs(depends on the converge rate) and lasts for about 6-10 hours. We add a different weight of the $r^{feats}$ for different configurations. The configuration $[w_{f_{0}^1}, w_{f_0^2}, w_{f_0^3}, w_{f_0^4}, w_{f_0^5}, w_{f_1}, w_{f_2}, w_{f_3}]$ represents the corresponding weight. 
The hyper-parameters can be found in Table~\ref{tab:tune}.

\begin{table}[ht]
\centering
\small
\begin{tabular}{c|c|c}
\toprule
\textbf{Feature Configurations} & \textbf{Reward Weight}& \textbf{Threshold} \\
\midrule
\text{[0, 1, 1, 1, 1, 0, 0, 0]} & 0.008&1 \\
\text{[1, 1, 1, 1, 1, 0, 0, 0]}& 0.008 &1\\
\text{[0, 1, 1, 1, 1, 0.5, 0, 0]}& 0.005 &1\\
\text{[1, 1, 1, 1, 1, 0.5, 0, 0]}& 0.005 &1\\
\text{[1, 1, 1, 1, 1, 0.5, 0.5, 0]}& 0.005&1 \\
\bottomrule
\end{tabular}
\caption{Hyper-parameters for Features Alignment}
\label{tab:tune}
\end{table}
\subsubsection{Active Strategy in Data Generation}
We employ active strategy in generating data as formulated below:
\begin{equation}
\begin{split}
        \text{W}_j = s_0 &+ w_{succ} \frac{\max_i{sr_i} - sr_j}{\max_i{sr_i} - \min_i{sr_i}} \\
        &+ w_{disc} \frac{\max_i{\overline{p_i}} - \overline{p_j}}{\max_i{\overline{p_i}} - \min_i{\overline{p_i}}}
\end{split}
\end{equation}
where $W_j$ denotes the overall score of the $j$-th task. In our implementation, we set $s_0, w_{succ}, w_{disc}$ to $0.2, 0.4, 0.4$, respectively.

\subsubsection{High-level Policy Training}

Using a task-specific reward \(r = w_{G} r_{G} + w_{p_1} r_{p_1} + w_{p_2} r_{p_2}\), we train a high-level policy to execute reaching and grasping. The training hyper-parameters are provided in Table~\ref{tab:highlevel}. Furthermore, the parameters related to the reward design are detailed in Table~\ref{tab:reward_params}.

\begin{table}[ht]
\centering
\small
\begin{tabular}{ll}
\toprule
\textbf{Hyper-parameters} & \textbf{Value} \\
\midrule
 {Learning Rate} & 2e-5 \\
 Episode Length & 300 \\
Action Distribution Variance & 0.1\\ 
 {Discount $\gamma$} & 0.99 \\
TD($\lambda$) & 0.95 \\
Disc Mini-batchsize & 4096\\

 {Policy Mini-batchsize} & 16384 \\
  {LR Schedule} & constant \\

 {Disc Grad Penalty Weight} & 5 \\

 {Disc Weight Decay} & 0.0001 \\

 $w_{G}$ & 0.4 \\

 $w_{p_1}$ & 0.2 \\

$w_{p_2}$ & 0.4 \\
\bottomrule
\end{tabular}
\caption{Hyper-parameters for High-level Policy Training}
\label{tab:highlevel}
\end{table}

\begin{table}[htbp]
\centering

\begin{tabular}{cc}
\toprule
\textbf{Parameter} & \textbf{Value} \\
\midrule
$w_{\text{pos}}$ & 0.3  \\
$w_{\text{vel}}$ & 0.6  \\
$w_{\text{face}}$ & 0.1\\
$w_{\text{grasp}}$ & 1.0  \\
$w_{\text{height}}$ & 2.0  \\
$w_{\text{obj-vel}}$ & 1.0\\
$h_{\text{lift\_target}}$ & 0.2 m\\
$v_{\text{threshold}}$ & 30m/s\\

\bottomrule
\end{tabular}
\caption{Reward Function Parameters}
\label{tab:reward_params}
\end{table}

\section{Experimental Details}
\label{secC}
\subsection{Details about Baselines}
In the implementation of AMP, we adopted a \textit{7:3} ratio between the task loss and discriminator loss, which produced the highest success rate during our experiments. When the ratio was set to \textit{1:0}, AMP degraded to Fullbody PPO. The same ratio was used for AMP*, with the only difference being that the motion prior in AMP* was trained on a dataset that included interpolated data, similar to Ours.

For PMP and PSE, we modified the discriminator in AMP to a part-wise design and trained it on walking data. In PMP, we adopted the same \textit{7:3} ratio between the task loss and discriminator loss as used in AMP. To further improve the success rate, we implemented a dedicated arm module for grasping, using rollout trajectories as references. During the second and third stages of the reaching task, the right-hand joints were encouraged to move closer to the reference. Similarly, in PSE, we introduced a part-wise discriminator as well. However, in the space training stage, the task and discriminator weight ratio was set to \textit{1:1}.

\subsection{Details about User Study}
In this study, human preference was used to evaluate the naturalness of motions. Specifically, we recruited \textbf{100} volunteers to compare the performance of different policies through video-based assessments. The evaluation process consisted of the following steps:
\begin{itemize}
    \item \textbf{Random Motion Generation}: For each policy, we randomly selected four sets of scene parameters and rendered motion sequences for these parameters in the Isaac simulator.
    \item \textbf{Side-by-Side Video Comparison}: For each pair of policies to be compared, the motion sequence videos were presented side by side to the volunteers. Each video was recorded from three different viewpoints, allowing the volunteers to comprehensively observe the motion performance.
    \item \textbf{Volunteer Judgments}: After viewing the videos, the volunteers selected the policy that they perceived to exhibit more natural motion.
\end{itemize}

The users will rate the motions from $N$ different aspects, selecting the best option for each aspect. The table in the main text shows the results of these comparisons. The baseline values represent the weighted sum of the proportion of volunteers who selected the baseline policy as the better option(in our case, the weights are all $\frac{1}{N}$).
\subsubsection{Users(Q): Quality Evaluation}
This evaluation focuses on the overall quality of the motions. For the grasping process, we further divided the assessment into four aspects: approaching the table, the grasping process, moving towards the target, and overall coherence.

\begin{itemize}
    \item \textbf{Naturalness of Approaching the Table}: This evaluates how the integration of additional data influences the naturalness of the original walking motion.
    \item \textbf{Grasping Process}: As a crucial part of the entire sequence, this examines how naturally the agent extends its hand and successfully grasps the object.
    \item \textbf{Moving Towards the Target}: This assesses the agent's ability to recover and move towards the target smoothly after grasping.
    \item \textbf{Overall Coherence}: This evaluates the naturalness of transitions between different stages of the motion and the overall intentionality of the entire sequence.
\end{itemize}
\subsubsection{Users(I): Issues Judgment}
In this evaluation criterion, we focus on the evaluation of detailed issues. Users are instructed to pay attention to specific details we identified (commonly observed unnatural patterns) and select the policy with fewer issues.

When comparing against the baseline, we highlighted four representative and classic issues: shuffling, sliding, near-loss of balance, and overly exaggerated or unnecessary movements.

For the ablation study, we first asked a subset of users to watch CIRCLE~\cite{circle} interpolated data and real MoCap data. From their feedback, we identified the most frequently mentioned issue keywords, which were then used as evaluation criteria. We provided users with six motions corresponding to a specific ratio and asked them to rank these motions from worst to best based on the criteria mentioned above. A motion ranked in the $i$-th position received a score of $i-1$. Finally, the overall scores are presented in the table.

\subsection{Details about Pred Score}
Considering that the visual differences introduced by the added features are relatively subtle, we adopted a more fine-grained evaluation approach for this section. Specifically, we used interpolated data generated by the CIRCLE strategy as negative examples and real motion capture data as positive examples to train a discriminator. This discriminator was then used to evaluate our generated motions. The scores presented in the table represent the average scores of 1,000 randomly sampled motions.

Specifically, we utilized the pre-trained feature extractor from MotionGPT to extract 512-dimensional features. Then these features were passed through three fully connected layers with a structure
[512, 128, 128, 1] per unit. After activation, the final output was the score. We retained the checkpoint with the best validation accuracy, achieving a discrimination accuracy of 94.2\%.



\section{Additional Experiments and Visualization}
\label{secD}




\subsection{Walking Phase Validation}

Our method generates natural, high-quality walking motions, especially when far from the table (full video will be provided instead of clipping around grasping). We evaluated the 1st-phase motion using a discriminator trained on walking MoCap, achieving disc rewards of 0.487 which is close to those of directly reproducing motions using ASE/AMP (0.503/0.516) and the oracle (0.492), compared to the real data rewards of 0.892. During the pre-grasping(2nd-phase), while our method demonstrates significant improvements over existing approaches like Omnigrasp and Braun's, some motions exhibit artifacts like sliding adjustments, primarily due to the need to generalize to \textbf{varying table widths} and the inherent complexity of interaction and \textbf{collision avoidance}. Even the oracle policy in mid-height shows noticeable adjustments in generalized scenarios. Noticing that MoCap data with high precision does not fully address these challenges and MoCap has its own limitations and biases, we turn to \textbf{explore more flexible synthetic data.} 

\subsection{Diversity of Motions}
\begin{figure}
    \centering
    \includegraphics[width=0.7\linewidth]{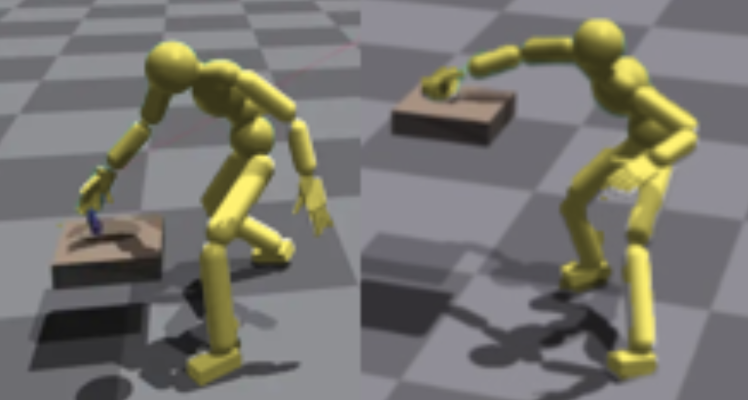}
    \caption{Diversity visualization}
    \label{div}
\end{figure}

\textbf{Diversity} was not a primary focus in our approach, as we prioritized maximizing SR during RL exploration and pose generation. This led to optimized poses for different table heights converging. Enhancing pose diversity and increasing variance during exploration could yield different grasping patterns(shown in Figure~\ref{div}).

\subsection{More about Feature Alignment}
In this section, we want to further analyze the effects of our feature alignment mechanism.
\begin{figure}
    \centering
    \includegraphics[width=0.8\linewidth]{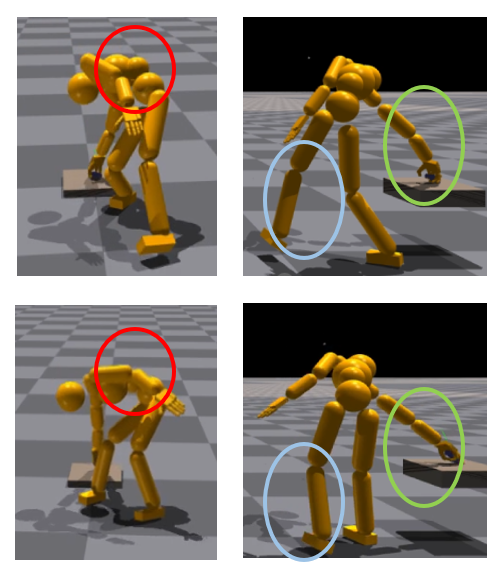}
    \caption{\textbf{Comparison of $f_0$ alignment:} The top image shows no alignment, while the bottom with  $f_0$ alignment reveals key changes: the torso bends at the lower joint (red circle), a passive reach becomes a coordinated motion driven by the upper arm (green circle), and a suspended leg transitions to a standing leg (blue circle).}
    \label{fig:pos-compare}
\end{figure}

\begin{figure*}
    \centering
    \includegraphics[width=0.9\linewidth]{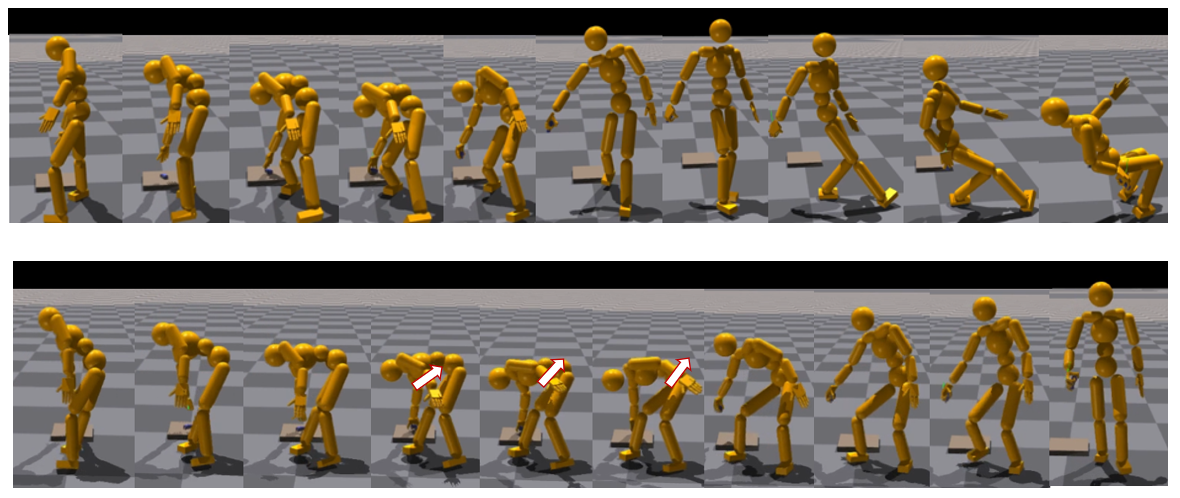}
    \caption{
\textbf{Visualization of Stability Enhancement through Feature Alignment:} The lower figure, with $f_0$ and $f_1$  alignment, significantly improves stability during both the \textit{grasp} phase and the \textit{recover to walk} stage. The left hand raises swiftly (indicated by the arrow), and the left foot steps back quickly to maintain balance when grasping low objects.}
    \label{fig:pos-compare2}
\end{figure*}

\textbf{First, the effect of adding a shallow layer is particularly evident in the \textit{partwise} patterns.} For example, as shown in the red circles in Figure~\ref{fig:pos-compare}, when we add the torso feature \(f_0^1\), the bending pattern of the torso changes. Initially, the bending originates from the closest joint to the head (torso1, corresponding to the chest and neck). 
After the addition, the upper body becomes mostly upright (borrowed from the \textit{walk} pattern), with the bending localized to torso2 (similar to the waist). On the other hand, as indicated by the green circles, the hand posture without alignment appears relatively relaxed. This frequently results in body tilt dominating the motion, where the upper arm remains stiff while the forearm dangles loosely in a \textit{reach} pattern. However, after alignment, a more natural pattern emerges: the upper arm drives the extension of the forearm, resulting in a more straight and coordinated reaching motion.

\textbf{Beyond the visual differences, it is even more critical to emphasize the contribution of this feature to overall stability.} Specifically, we observe that this feature significantly aids both the final stage of \textit{grasp} and the \textit{recover to walk} phase. For example(as shown in Figure~\ref{fig:pos-compare2}), with the addition of features(especially $f_1$), the motion exhibits a clear pattern where the left hand is swiftly raised and the left foot quickly steps back to maintain balance when grasping very low objects. This coordinated movement of the limb is particularly beneficial for dynamic balance during the recovery phase in \textit{walk}.

We observe that without the alignment of the features, the agent adopts a more 'aggressive' posture characterized by a bent right leg and a lifted left foot. While this posture is inspired by real grasping motions (e.g. actions similar to picking up a small ball in the ground), these movements rely on fine muscle coordination and balance that are challenging even for humans. Such extreme low-grasp postures make it difficult to recover quickly, let alone maintain balance for the agent. 

To achieve the same task, a more robust strategy can be adopted to enhance the success rates. For example, keep both feet grounded and squat sideways instead of tilting. This movement resembles natural upright postures and walking, which inherently favor balanced grasping. \textbf{This strategy of “learning more stable task-completion motions” from \textit{walk} can be further incorporated into humanoid robotics to help execute tasks.}



We alse observed that \textbf{incorporating deeper features}, such as $f_2$ and $f_3$, during feature alignment can \textbf{negatively impact the overall learning of reaching skills}. Specifically, when $f_2$ is included, the success rate decreases, even falling below the performance achieved without additional data. Furthermore, when $f_3$ is added, the policy fails to successfully walk to the table. This decline in performance arises because deeper features capture more information about global movement(as shown in the pilot study below). For new datasets, the objective of reaching the target and the goal of maintaining local features close to those from the walking diverge significantly. This divergence introduces inconsistency into the training process, leading to a disorganized feature space that hinders effective learning of the task.

\subsection{Ablation Study of Data Ratio}

In our active strategy, we emphasize \textbf{maximizing the utility of generated data to achieve high success rates with a minimal data ratio}. In this section, we provide a detailed analysis of the impact of the data ratio on policy performance and the role of "maximizing the utility of generated data."

\begin{figure}[!ht]
    \centering
    \begin{subfigure}{0.3\textwidth}
        \centering
        \includegraphics[width=\linewidth]{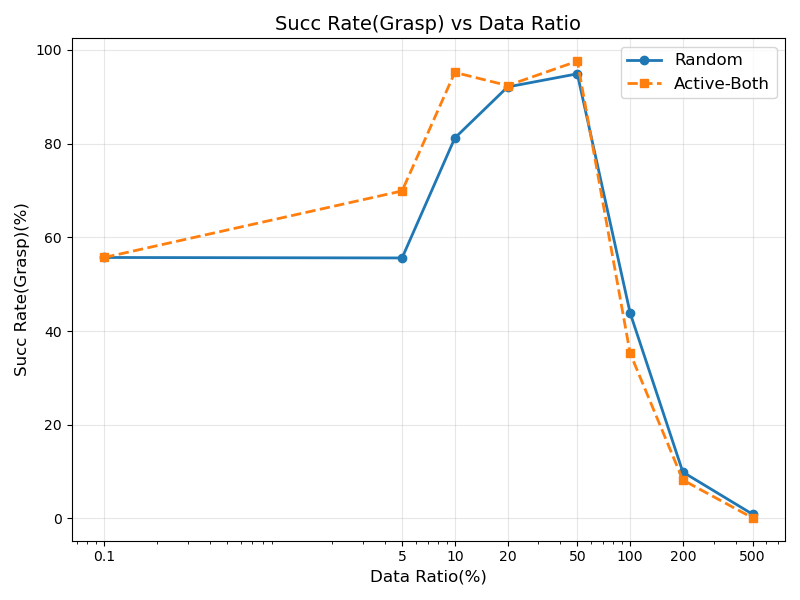}
    \end{subfigure}
    \hspace{-2mm}
    \begin{subfigure}{0.3\textwidth}
        \centering
\includegraphics[width=\linewidth]{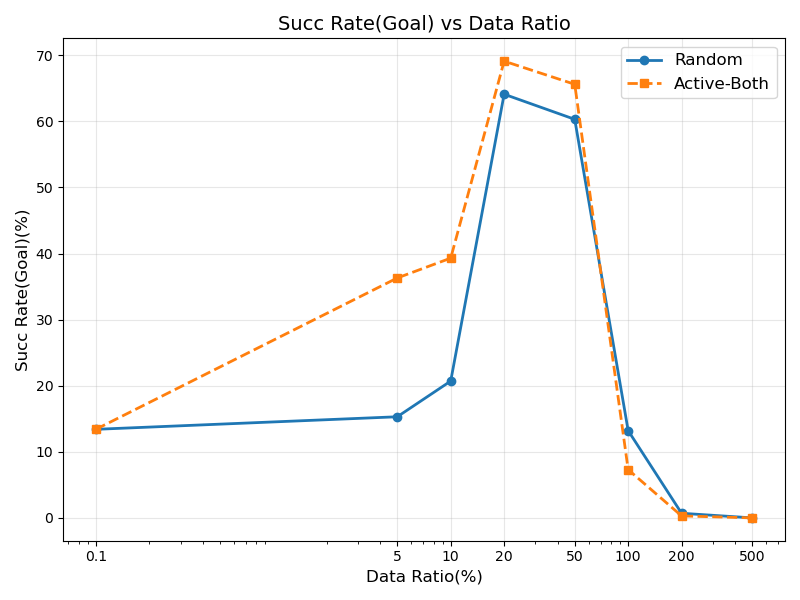}
    \end{subfigure}
    \caption{\textbf{Success Rate Curve with Varying Data Ratio(\%)}: Using a small data ratio allows for an active strategy that specifically targets challenging tasks, resulting in better success rate. However, an excessively large data ratio can lead to a rapid decline in the success rate.}
    \vspace{-2mm}
    \label{curve_succ}
\end{figure}

As previously defined, the data ratio refers to the percentage of generated data relative to the original data used during sampling. As shown in Figure~\ref{curve_succ}, we observe that when the data ratio is low, the success rate increases rapidly as the ratio grows, peaking at around 20-30\%. In this phase, the active strategy plays a crucial role by specifically addressing tasks that walk data cannot handle, effectively avoiding the addition of irrelevant data.

\begin{figure*}[!ht]
    \centering
    \small
    \includegraphics[width=\linewidth]{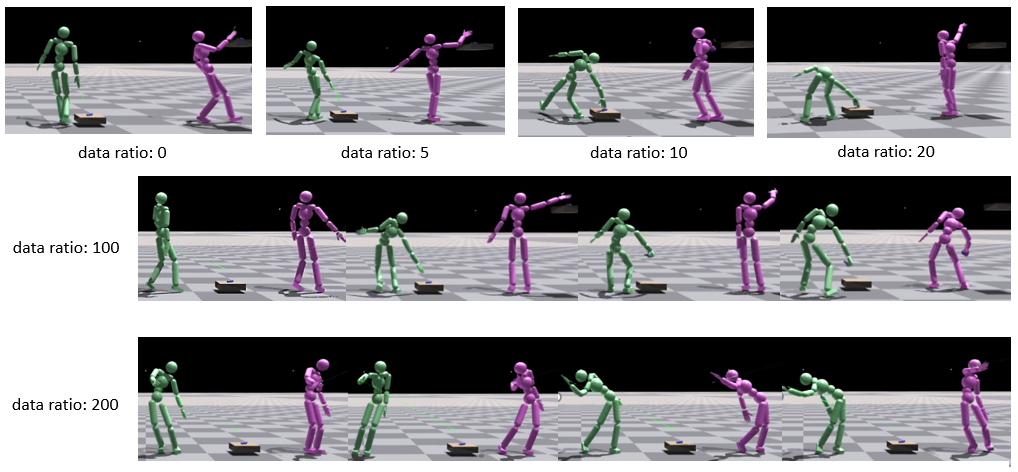}
    \caption{\textbf{Visualization of Reaching and Grasping with Varying Data Ratios(\%):} At low data ratios, task completion improves rapidly as the ratio increases. However, when the ratio exceeds 100\%, the character struggles with natural turning, and beyond 200\%, the character shifts focus to balancing between skills, hindering effective walking.}
    \label{ratio}
\end{figure*}

However, adding more data does not necessarily compensate for the shortcomings of the random strategy. As we can observed, the success rate drops sharply when the ratio continuously increase, reaching nearly zero when the ratio exceeds 1:1. This is because the excessive addition of data makes learning the skill space more challenging. As new data increases, natural walking and turning skills are forgotten. As shown in the Figure~\ref{ratio}, when the ratio exceeds 1:1, the character struggles to perform natural turns. When the ratio exceeds 1:2, the character focuses on switching between various skills to maintain balance, making it difficult to walk effectively.

Even with a ratio below 1:1 (data ratio $\leq 100$), we observe that larger ratios result in significantly longer training times. As seen in ASE~\cite{ASE}, even well-balanced weight designs require several days of training for relatively small differences, such as a 30-minute skill. This further highlights the value of first training a fundamental walk space then selectively expanding it. By doing so, we efficiently learn the truly reusable "walk" skill and focus on the task-relevant "reach" skill at minimal cost.

\section{More about Pilot Study}
\label{secF}
\subsection{Pilot study with larger dataset}
In our pilot study, features from shallow layers show distinct clustering in real data and lower FID values. In contrast, deeper layers prioritize semantic information, reducing this clustering. However, even in deep layers, the FID values between MoCap Reach and MoCap Walk are lower than those between MoCap Reach and Generated Reach.

We hypothesize that this discrepancy arises due to the limited training data: although MoCap Reach and Generated Reach share similar global movements (e.g. reaching), the critic primarily identifies that they are distinct from "walk-like" motions but lacks a comprehensive understanding of the specific characteristics of reaching motions.

We believe that with a critic trained on a larger and more diverse dataset, MoCap Reach and Generated Reach—both representing similar semantic motions—would cluster more closely in the deep layers. This would further support the notion of a universal phenomenon: transferable local patterns are captured in shallow layers, whereas deeper layers reflect global, task-specific movements.
\subsubsection{Dataset}

To enhance the critic's ability to understand motions, we use more data to train the critic. The added data can be find below in Table~\ref{tab:pilot_study_plus}:
\begin{table}[htbp]
\centering
\small
\begin{tabular}{ll}
\toprule
\textbf{File Name} & \textbf{Weight} \\
\midrule
RL\_Avatar\_Atk\_Spin\_Motion.npy & 0.00724638 \\
RL\_Avatar\_Standoff\_Feint\_Motion.npy & 0.03105590 \\
RL\_Avatar\_Dodge\_Backward\_Motion.npy & 0.01552795 \\
RL\_Avatar\_RunBackward\_Motion.npy & 0.01552795 \\
RL\_Avatar\_WalkBackward01\_Motion.npy & 0.01552795 \\
RL\_Avatar\_WalkBackward02\_Motion.npy & 0.01552795 \\
RL\_Avatar\_Dodgle\_Left\_Motion.npy & 0.01552795 \\
RL\_Avatar\_RunLeft\_Motion.npy & 0.01552795 \\
RL\_Avatar\_WalkLeft01\_Motion.npy & 0.01552795 \\
RL\_Avatar\_WalkLeft02\_Motion.npy & 0.01552795 \\
RL\_Avatar\_Dodgle\_Right\_Motion.npy & 0.01552795 \\
RL\_Avatar\_RunRight\_Motion.npy & 0.01552795 \\
RL\_Avatar\_WalkRight01\_Motion.npy & 0.01552795 \\
RL\_Avatar\_WalkRight02\_Motion.npy & 0.01552795 \\
RL\_Avatar\_RunForward\_Motion.npy & 0.02070393 \\
RL\_Avatar\_WalkForward01\_Motion.npy & 0.02070393 \\
RL\_Avatar\_WalkForward02\_Motion.npy & 0.02070393 \\
RL\_Avatar\_Standoff\_Circle\_Motion.npy & 0.06211180 \\
RL\_Avatar\_TurnLeft90\_Motion.npy & 0.03105590 \\
RL\_Avatar\_TurnLeft180\_Motion.npy & 0.03105590 \\
RL\_Avatar\_TurnRight90\_Motion.npy & 0.03105590 \\
RL\_Avatar\_TurnRight180\_Motion.npy & 0.03105590 \\
RL\_Avatar\_Fall\_Backward\_Motion.npy & 0.00869565 \\
RL\_Avatar\_Fall\_Left\_Motion.npy & 0.00869565 \\
RL\_Avatar\_Fall\_Right\_Motion.npy & 0.00869565 \\
RL\_Avatar\_Fall\_SpinLeft\_Motion.npy & 0.00869565 \\
RL\_Avatar\_Fall\_SpinRight\_Motion.npy & 0.00869565 \\
RL\_Avatar\_Idle\_Alert(0)\_Motion.npy & 0.00434783 \\
RL\_Avatar\_Idle\_Alert\_Motion.npy & 0.00434783 \\
RL\_Avatar\_Idle\_Battle(0)\_Motion.npy & 0.00434783 \\
RL\_Avatar\_Idle\_Battle\_Motion.npy & 0.00434783 \\
RL\_Avatar\_Idle\_Ready(0)\_Motion.npy & 0.00434783 \\
RL\_Avatar\_Idle\_Ready\_Motion.npy & 0.00434783 \\
CMU\_07\_02.npy & 0.04070393 \\
CMU\_07\_01.npy & 0.04070393 \\
CMU\_07\_07.npy & 0.04070393 \\
amp\_humanoid\_jog.npy & 0.08316768 \\
amp\_humanoid\_walk.npy & 0.09316768 \\
amp\_humanoid\_run.npy & 0.08316768 \\
\bottomrule
\end{tabular}
\caption{\textbf{Critic Dataset}: Weights for different motions}
\label{tab:pilot_study_plus}
\end{table}

\subsubsection{Result}
\begin{figure*}[!ht]
    \centering
    \begin{subfigure}{0.26\textwidth}
        \centering
        \includegraphics[width=\linewidth]{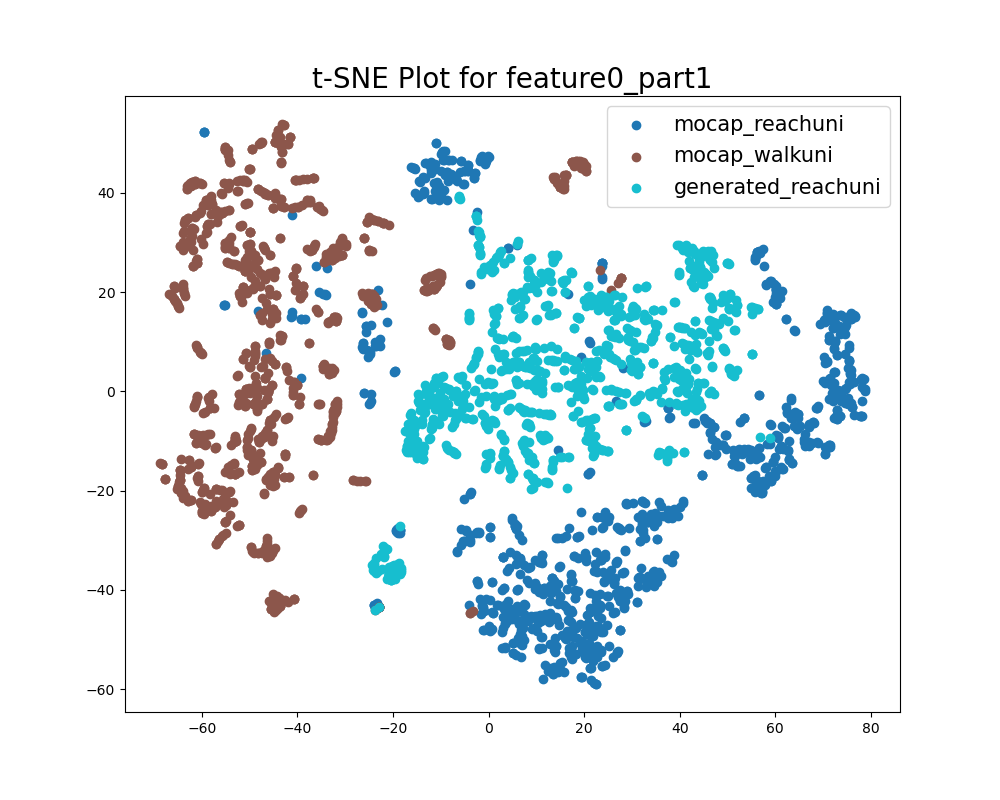}
    \end{subfigure}
    \hspace{-6mm}
    \begin{subfigure}{0.26\textwidth}
        \centering
\includegraphics[width=\linewidth]{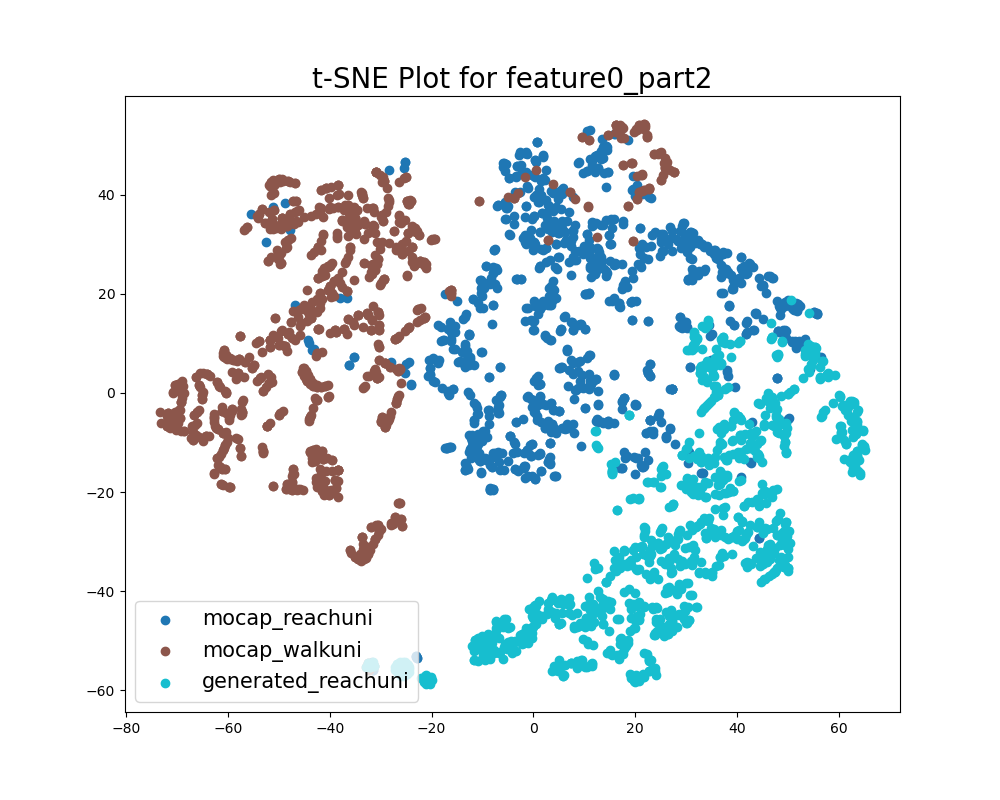}
    \end{subfigure}
        \hspace{-5.5mm}
    \begin{subfigure}{0.26\textwidth}
        \centering
        \includegraphics[width=\linewidth]{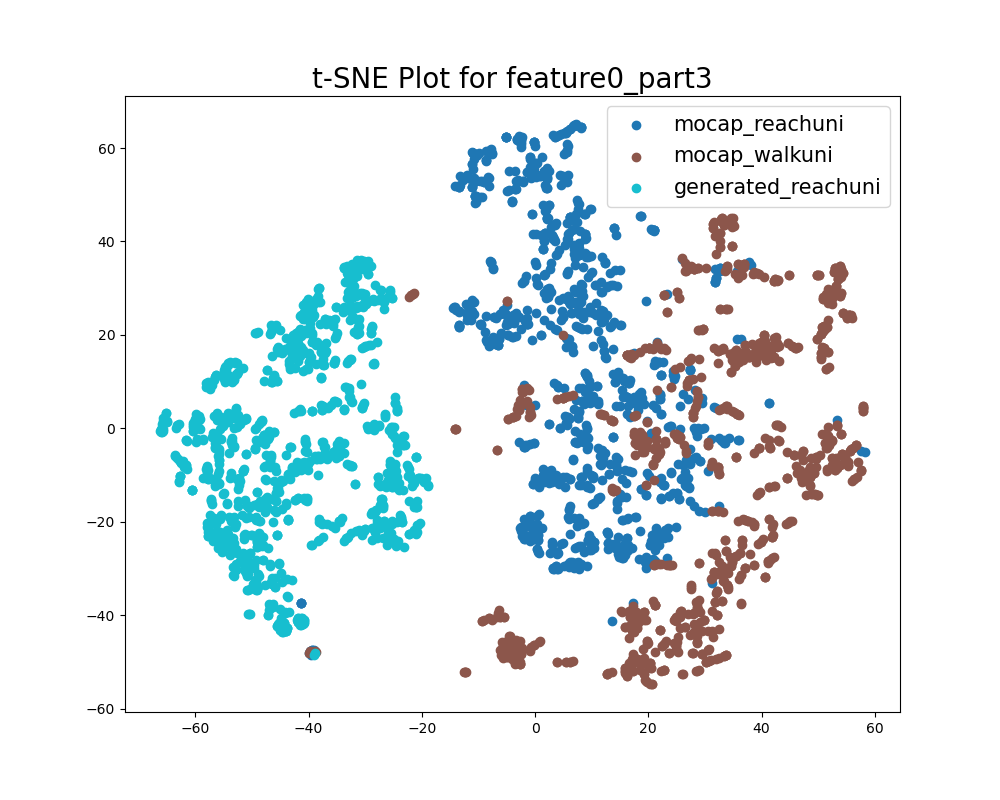}
    \end{subfigure}
        \hspace{-5.5mm}
    \begin{subfigure}{0.26\textwidth}
        \centering
        \includegraphics[width=\linewidth]{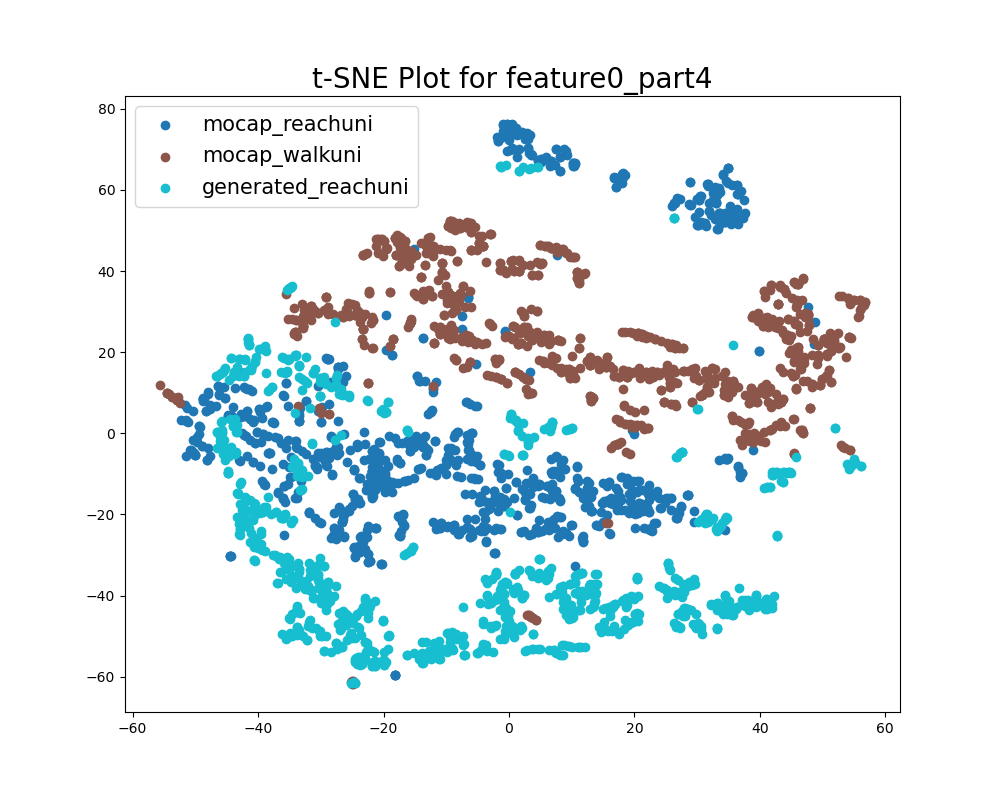}
    \end{subfigure}
    \begin{subfigure}{0.26\textwidth}
        \centering
        \includegraphics[width=\linewidth]{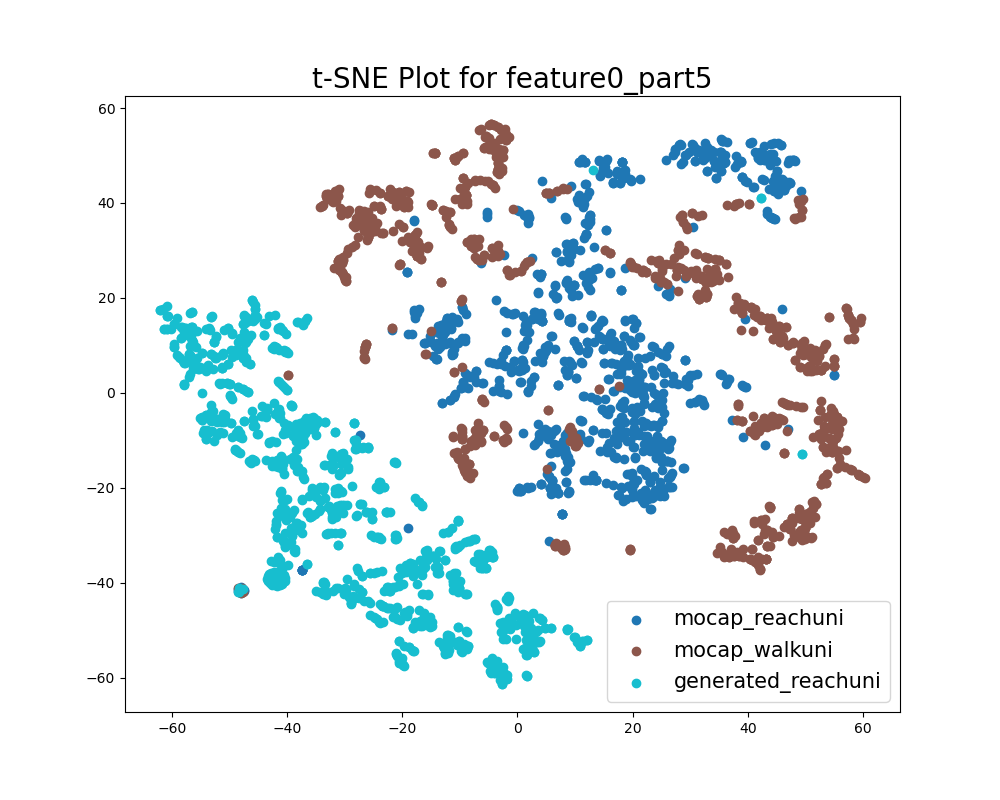}
    \end{subfigure}
          \hspace{-5.5mm}
    \begin{subfigure}{0.26\textwidth}
        \centering
        \includegraphics[width=\linewidth]{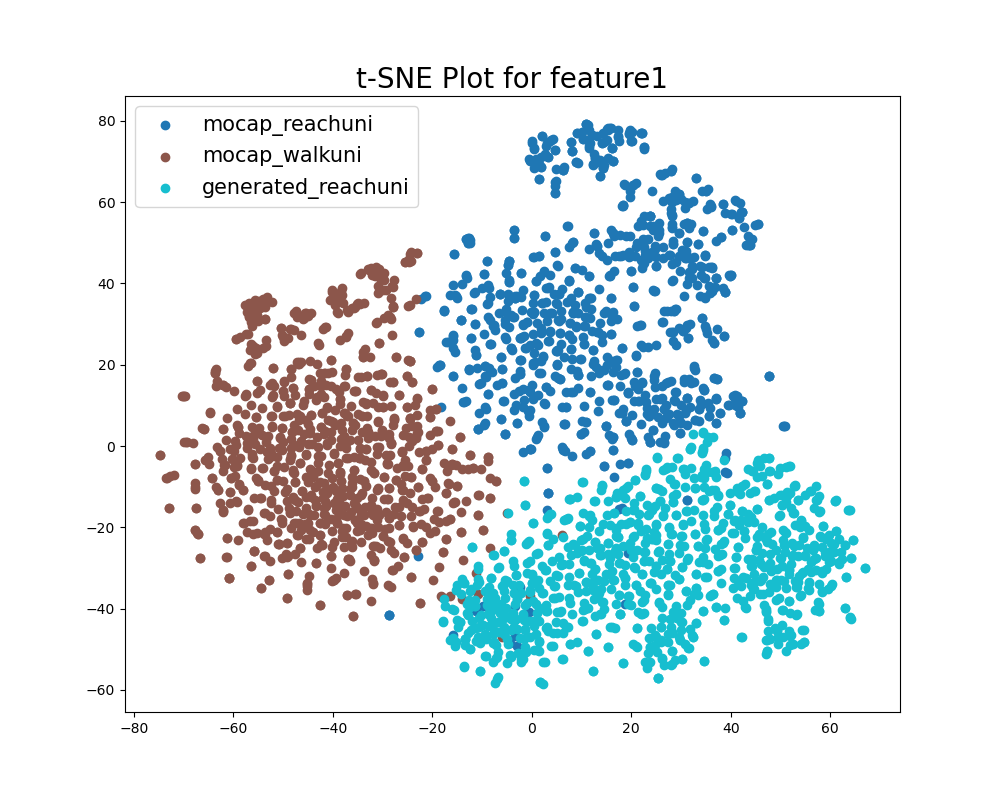}
    \end{subfigure}
         \hspace{-5.5mm}
    \begin{subfigure}{0.26\textwidth}
        \centering
        \includegraphics[width=\linewidth]{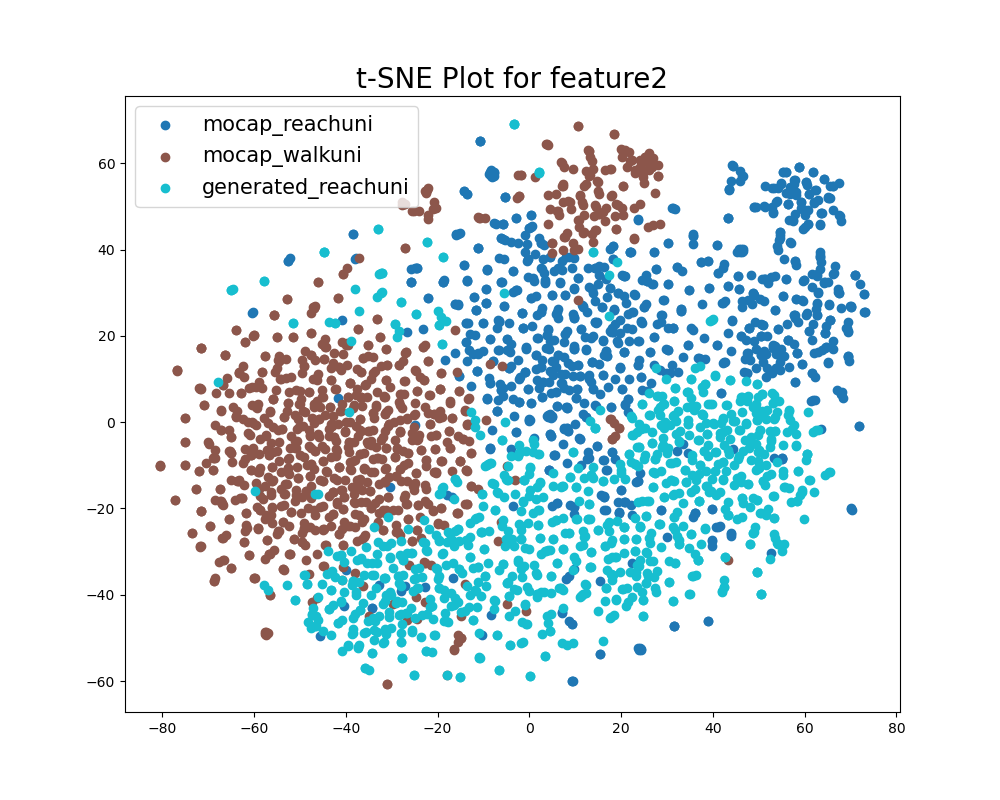}
    \end{subfigure}
           \hspace{-5.5mm}
    \begin{subfigure}{0.26\textwidth}
        \centering
        \includegraphics[width=\linewidth]{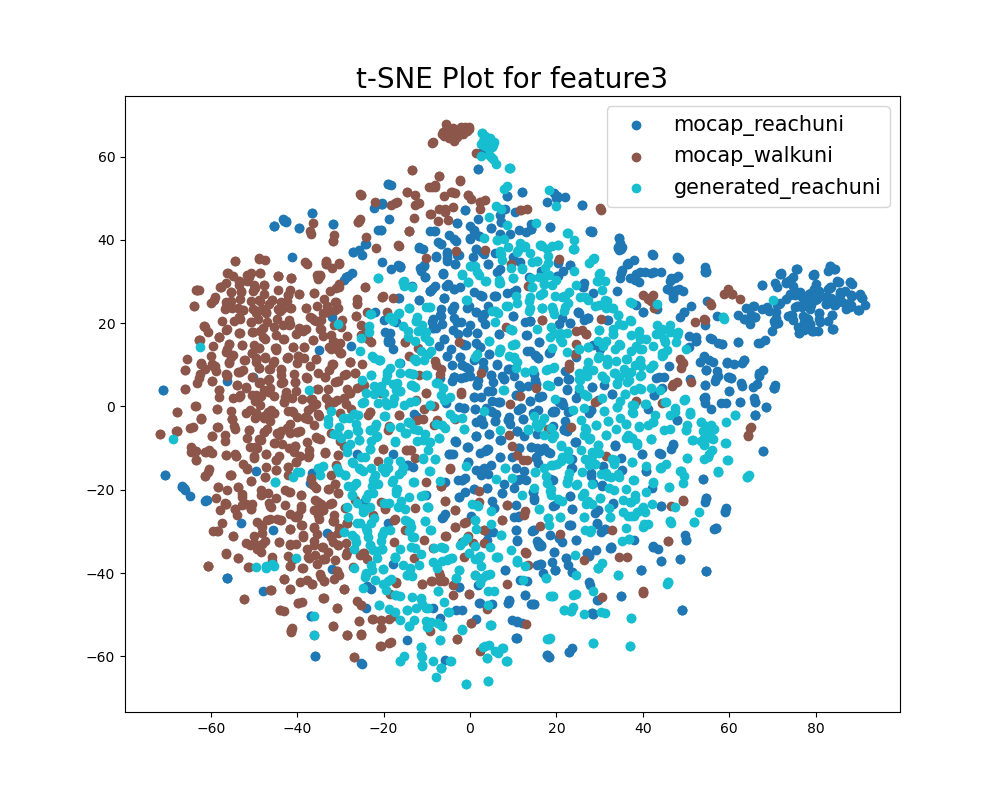}
    \end{subfigure}
    \caption{\textbf{t-SNE plots of features extracted at different levels of the more comprehensive critic network}: There is clear clustering within the MoCap data in shallow layers and a clear clustering within Reach data in deeper layers.}
    \vspace{-2mm}
    \label{fig:pilot_study_plus}
\end{figure*}

When using a critic trained with a larger dataset, we observed deeper-level clustering of two macroscopic motion types, particularly in the final two layers of the network. At the same time, shallow-level clustering remained similar to that observed with our previous critic(shown in Figure~\ref{fig:pilot_study_plus}). Although our dataset does not contain specific reaching motions, the increased data volume enhanced the critic's understanding. As a result, the value predictions were no longer solely based on "walk-like" patterns but also incorporated the classification of macroscopic motion types.

Further pilot studies revealed that, under our critic network architecture, shallow-level features effectively capture the patterns of real motion and can transfer across different motion types and tasks. In contrast, deeper-level features encode more information about macroscopic motion forms and semantic details.

\subsection{Pilot Study with various motions}
\begin{figure}[ht]
    \centering
    \includegraphics[width=\linewidth]{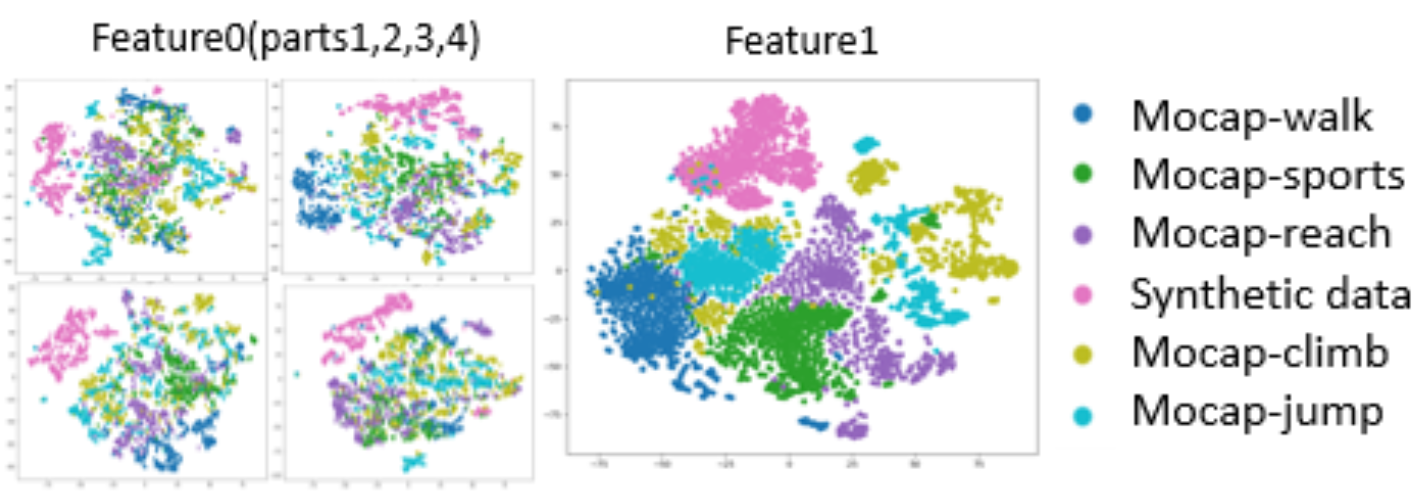}
    \caption{pilot study for various task}
    \label{pilottask}
\end{figure}

Reaching and grasping are fundamental yet highly challenging tasks that require balancing high DOF while enabling precise manipulation, making them a rigorous testbed for motion generation. We also conducted pilot studies across various tasks , obtaining consistent results. All these motions are from AMASS~\cite{AMASS}. For tasks beyond reaching, like shown in the Figure~\ref{pilottask}, MoCap data cluster together in shallow layers while synthetic data remain separated. These suggest our mechanism can capture transferable patterns from walking beyond specific tasks.

\end{document}